\documentclass[a4paper,fleqn]{cas-dc}
\usepackage{algorithm}
\usepackage{algpseudocode}
\usepackage{amsmath, amssymb} 
\usepackage{graphicx}
\usepackage{amsfonts}
\usepackage{array}
\usepackage{caption}
\usepackage{multirow}
\usepackage{booktabs}
\usepackage{subcaption}
\usepackage{float}
\usepackage[T1]{fontenc} 
\usepackage{pgf-pie}
\usepackage[utf8]{inputenc} 
\usepackage{multirow}
\usepackage{caption}
\usepackage{textcomp}
\usepackage{geometry}
\usepackage{hyperref}
\usepackage{tikz}
\usepackage{adjustbox}
\usepackage{booktabs}
\usepackage{amsmath}
\usepackage{bm}
\usepackage{tabularx}
\usepackage[numbers,sort&compress]{natbib}
\usepackage{flushend}
\usepackage{xcolor}
\usepackage[nameinlink,capitalize]{cleveref}
\hypersetup{colorlinks=true, linkcolor=blue, filecolor=cyan, citecolor=blue, urlcolor=blue}
\usepackage[short labels]{enumitem}
\usepackage[english] {babel}
\usepackage{pifont}
\addto\extrasenglish{

}

\def\tsc#1{\csdef{#1}{\textsc{\lowercase{#1}}\xspace}}
\tsc{WGM}
\tsc{QE}
\begin{document}
\let\WriteBookmarks\relax
\def\floatpagepagefraction{1}
\def\textpagefraction{.001}

\newcommand{\bluecheck}{}%
\DeclareRobustCommand{\bluecheck}{%
  \tikz\fill[scale=0.4, color=blue]
  (0,.35) -- (.25,0) -- (1,.7) -- (.25,.15) -- cycle;%
}
\newcommand{\cmark}{\ding{51}}%
\newcommand{\xmark}{\ding{55}}%

    
   
    
   

 \shorttitle{}
\shortauthors{Dar \textit{et al.} }


\title [mode = title]{An Explainable Deep Neural Network with Frequency-Aware Channel and Spatial Refinement for Flood Prediction in Sustainable Cities}

\author[1]{Shahid Shafi Dar}[auid=1]
\ead{phd2201201004@iiti.ac.in}

\author[1]{Bharat Kaurav}
\ead{cse220001017@iiti.ac.in}

\author[1]{Arnav Jain}
\ead{ee220002018@iiti.ac.in}
\author[1]{Chandravardhan Singh Raghaw}
\ead{phd2201101016@iiti.ac.in}
\author[1]{Mohammad Zia Ur Rehman}
\ead{phd2101201005@iiti.ac.in}

\author[1]{Nagendra Kumar}
\cormark[1]
\ead{nagendra@iiti.ac.in}
\cortext[cor1]{Corresponding author}

\affiliation[1]{organization={Indian Institute of Technology Indore}, postcode={453552}, country=India}
\begin{abstract}
In an era of escalating climate change, urban flooding has emerged as a critical challenge for sustainable cities, threatening lives, infrastructure, and ecosystems. Traditional flood detection methods are constrained by their reliance on unimodal data and static rule-based systems, which fail to capture the dynamic, non-linear relationships inherent in flood events. Furthermore, existing attention mechanisms and ensemble learning approaches exhibit limitations in hierarchical refinement, cross-modal feature integration, and adaptability to noisy or unstructured environments, resulting in suboptimal flood classification performance.
To address these challenges, we present XFloodNet, a novel framework that redefines urban flood classification through advanced deep-learning techniques. XFloodNet integrates three novel components: (1) a Hierarchical Cross-Modal Gated Attention mechanism that dynamically aligns visual and textual features, enabling precise multi-granularity interactions and resolving contextual ambiguities; (2) a Heterogeneous Convolutional Adaptive Multi-Scale Attention module, which leverages frequency-enhanced channel attention and frequency-modulated spatial attention to extract and prioritize discriminative flood-related features across spectral and spatial domains; and (3) a Cascading Convolutional Transformer Feature Refinement technique that harmonizes hierarchical features through adaptive scaling and cascading operations, ensuring robust and noise-resistant flood detection. We evaluate our proposed method on three benchmark datasets, such as Chennai Floods, Rhine18 Floods, and Harz17 Floods, XFloodNet achieves state-of-the-art F1-scores of 93.33\%, 82.24\%, and 88.60\%, respectively, surpassing existing methods by significant margins.
\end{abstract}

\maketitle

\section{Introduction}

Floods rank among the most catastrophic natural disasters, disrupting lives, infrastructure, and ecosystems globally \citep{INCIDENTS, Priest}. Their sudden onset and widespread impact demand rapid assessment and response to mitigate damage and support recovery efforts \citep{Wang_2023}. Floods result in global economic losses exceeding \$40 billion annually\footnote{\href{https://wmo.int/topics/floods}{World Meteorological Organization}},  with urban areas bearing a significant brunt due to dense populations and critical infrastructure. In 2023 alone, approximately 32 million people were affected by flooding, experiencing injuries or displacement\footnote{\href{https://www.statista.com/statistics/1293353/global-number-of-people-affected-by-floods/}{Statista}}. As climate change intensifies, the frequency and severity of urban flooding are expected to rise, posing a growing threat to sustainable cities. Effective flood prediction and mitigation strategies are therefore crucial to minimize socio-economic and environmental impacts, ensuring the resilience of urban communities and ecosystems. \\
Deep learning techniques offer promising solutions for rapid and accurate flood detection, enabling cities to respond proactively to flood events \citep{social, social1}. The emergence of social media platforms have emerged as a transformative tool for disaster management, offering real-time, crowd-sourced data that can enhance situational awareness and decision-making during flood events. Social media platforms such as Twitter (now X) generate vast amounts of user-generated content, including images, videos, and text, which provide valuable insights into on-the-ground conditions \citep{Paul}. This data is particularly useful for flood classification due to its geographical coverage, temporal immediacy, and rich contextual information. For instance, images shared on social media can visually confirm flood occurrences, while textual posts often describe the severity of flooding. However, leveraging social media data for flood classification presents significant challenges. The noisy and unstructured nature of user-generated content such as low-quality images, informal language, and irrelevant posts requires advanced techniques to extract meaningful information. Additionally, the ambiguity of contextual information and the heterogeneity of data modalities complicate the integration of social media data into flood prediction models. Despite these challenges, the potential of social media data for flood classification is immense, particularly when combined with advanced deep learning techniques.

\subsection{Technical Challenges and Research Gaps}
Traditional methods \citep{postfloodcbam, bashir2024efficient, yasi2024flood} often rely on unimodal analysis, such as analyzing images or text in isolation, limiting their ability to capture the full context of flood events. This limitation necessitates the development of multimodal frameworks that can effectively combine image and text data for a more comprehensive understanding of flood events.\\
Efraimidou et al. \citep{Efraimidou_2024} proposed a deterministic, non-machine learning approach for flood risk assessment. However, this static method lacks adaptability to dynamic flood events and struggles with capturing complex, non-linear relationships. In contrast, XFloodNet employs a data-driven deep learning approach to learn these intricate relationships from multimodal data, overcoming the limitations of rule-based systems and providing more accurate and adaptable flood predictions. Bentivoglio et al. \citep{Bentivoglio_2022} demonstrated the effectiveness of Convolutional Neural Networks (CNNs) for flood mapping. However, CNNs alone may not fully capture the complex, multifaceted nature, especially in noisy environments \citep{CNNs_ViT}. To this end, XFloodNet addresses this by incorporating advanced attention mechanisms into its multimodal framework, which allows the model to selectively focus on relevant features, capturing fine-grained details and improving robustness.
Ellora et al. \citep{yasi2024flood} demonstrated the effectiveness of ensemble learning for parameter optimization in flood detection. However, ensemble methods face challenges like diminishing returns, overfitting, and difficulties in combining model outputs. XFloodNet tackles these challenges by employing a multimodal framework that combines joint image-text representations with a Heterogeneous Convolutional Adaptive Multi-scale Attention Module. This approach captures both point-wise and group-wise heterogeneous features, along with multi-scale information, offering a more efficient and comprehensive solution for flood classification.\\
Traditional attention mechanisms such as Convolutional Block Attention Module (CBAM) \citep{CBAM}, Efficient Channel Attention (ECA) \citep{ECA}, and Multimodal Channel Attention (MCA) \citep{dmcc} also exhibit limitations in handling the complexities of multimodal data. While MCA focuses on channel-level weighting within fused image and text modalities, and Cross-Attention Multimodal facilitates cross-modal interactions, both approaches fall short in several key areas. These methods often lack hierarchical refinement, fail to capture cross-modal interactions at a deeper semantic level, and are limited in their ability to extract spatial and frequency-domain features. Moreover, they struggle with adaptability to noisy or unstructured data. XFloodNet addresses these shortcomings by incorporating the Heterogeneous Convolutional Adaptive Multi-scale Attention and Hierarchical Cross-Modal Gated Attention mechanism, which enables more precise and robust multimodal learning, ensuring improved performance in diverse and challenging flood scenarios.
\subsection{XFloodNet: A Multimodal Framework for Urban Flood Resilience}
To address the aforementioned gaps in the literature, we propose XFloodNet, a novel multimodal framework designed to enhance urban flood classification through advanced deep-learning techniques. Our objective is to develop a high-performance flood classification method that effectively categorizes flood events by leveraging the information available in social media data. At the core of XFloodNet lies the Multimodal Feature Interaction Module, which learns joint image-text representations by employing hierarchical cross-modal gated attention. This mechanism facilitates the fusion of visual and textual features and resolves contextual ambiguities. XFloodNet incorporates a Heterogeneous Convolutional Adaptive Multi-scale Attention Module that combines Groupwise and pointwise convolutions to enhance feature extraction while maintaining gradient flow for effective learning. The features are refined using Frequency-Enhanced Efficient Channel Attention and Frequency-Modulated Spatial Attention, which operate across multiple scales. This enables the model to capture fine-grained and global contextual information, identify key regions, and adapt features across different frequency bands. By incorporating frequency-domain features in spatial and channel dimensions, XFloodNet effectively distinguishes flood indicators from background noise, improving robustness and accuracy under diverse and challenging conditions. Furthermore, Cascading Convolutional Transformer Feature Refinement Module employs a hierarchical feature extraction approach with progressive refinement to enhance classification performance. These interconnected modules enable XFloodNet to integrate spatial and contextual dependencies for flood classification effectively. Beyond technical excellence, XFloodNet's interpretable architecture fosters transparency and trust, empowering urban planners, policymakers, and emergency responders to make informed, data-driven decisions. By transforming complex, multimodal data into actionable insights, this framework directly supports the United Nations’ Sustainable Development Goal 11, enabling cities to build resilience against climate-induced disasters. XFloodNet not only advances flood prediction but also serves as a transformative approach for creating safer, more sustainable urban environments, driving meaningful societal impact in the face of unprecedented environmental challenges.

\subsection{Key Contributions of Proposed Approach}
This work makes several key contributions to flood classification:
\begin{enumerate}

    \item We propose XFloodNet, an innovative multimodal framework that integrates frequency domain features with attention mechanisms to align textual and visual features effectively. This alignment ensures robust and accurate flood classification by leveraging complementary information from multiple modalities.

    \item We propose a Hierarchical Cross-Modal Gated Attention mechanism that dynamically adjusts attention weights across multiple levels. This mechanism captures both intra-modal and inter-modal relationships, optimizing image-text interaction and enhancing cross-modal feature fusion. The result is a significant improvement in flood classification performance.

    \item We propose Frequency-Enhanced Efficient Channel Attention integrates high and low-frequency features within a single-channel framework. This integration enhances feature representation while maintaining computational efficiency, ensuring that the model can process complex flood imagery without excessive computational overhead.
    
    \item We introduce a Frequency-Modulated Spatial Attention mechanism that combines multi-scale convolutions with frequency domain features. This dual-domain approach enhances the model’s focus on spatially relevant regions and global patterns, leading to improving flood classification performance. 

    \item  We propose the Cascading Convolutional Transformer Feature Refinement Module (CCTFRM), which leverages gated convolutional networks and transformers with cascading operations for progressive feature refinement. This module enhances pattern recognition and classification performance in flood imagery by iteratively refining features at multiple levels.
     
    \item We propose the Reverse Attention Harmonization technique, which employs gated subtraction and adaptive scaling operations guided by trainable weights. This technique optimizes feature interaction between visual features and the output of CCTFRM, enhancing information flow and harmonizing refined features with visual features. The result is improved flood detection performance.

    \item We evaluate XFloodNet on real-world flood datasets, demonstrating its superior performance over state-of-the-art methods. The results highlight the framework's effectiveness in precisely extracting relevant flood-related features, underscoring its potential for practical applications in flood monitoring and response.
    
\end{enumerate}

\subsection{Outline of the Manuscript}
The remainder of this paper is organized as follows: \autoref{sec2} presents related works. \autoref{section3} describes problem statement. \autoref{sec3} elaborates on the methodology and the proposed architecture. \autoref{sec4} presents experimental results and comparative evaluations. \autoref{sec5} discusses challenges and future directions, followed by the conclusion in \autoref{sec6}.

\section{Related Work} \label{sec2} 
Flood classification has seen significant advancements with the integration of Deep Learning techniques, particularly those leveraging multimodal information. This section provides an overview of related work, categorized into two key areas: Multimodal Disaster Content Classification and Attention Mechanisms.
\subsection{ Multimodal Disaster Content Classification}
Multimodal methods have achieved significant milestones by integrating visual and textual information, demonstrating their utility in applications such as image classification, text classification, and retrieval \citep{alphaclip, vlm1, vlm2, vlm3, vlm4}. Various models such as CLIP \citep{CLIP} and ViLT \citep{vilt} excel in aligning visual features with semantic text embeddings, enabling robust generalization across diverse domains. Frameworks such as DeepUnseen \citep{10195016} have further demonstrated the efficacy of integrating classification and segmentation modules for disaster scene analysis. Leveraging CLIP-based masked image processing and hierarchical task structuring, DeepUnseen refines semantic predictions for degraded imagery under adverse conditions. These efforts underline the necessity of combining multimodal reasoning with specialized architectures to tackle the unique challenges of disaster and weather-related tasks.\\
Multimodal methods have advanced the disaster domain by leveraging the complementary strengths of both modalities. Pretrained multimodal models such as VisualBERT effectively align textual and visual features, capturing cross-modal dependencies that enhance contextual understanding and classification performance \citep{sirbu, CLIP4STR}. It has been observed that multimodal frameworks consistently outperform unimodal methods. These approaches are crucial for enhancing the accuracy and reliability of crisis-related content analysis \citep{madichetty}.
CrisisSpot's \citep{Dar} graph-based neural network captures subtle multimodal relationships, achieving state-of-the-art F1-scores on disaster datasets.  In low-data circumstances, multimodal approaches \citep{CLMC, mm1, mm2, mm3} with pre-trained models such as  CLIP \citep{CLIP} and ALIGN \citep{ALIGN} have shown high performance with minimal fine-tuning. Previous multimodal approaches have made considerable advancements in tackling the challenges of disaster content classification. However, many of these methods struggle to effectively capture and integrate complex features from diverse modalities. They often fail to account for both intra-modal and inter-modal interactions, which are essential for improving classification performance. Furthermore, these approaches neglect the use of multi-scale convolutional operations, which are crucial for addressing image variability in social media data.
\subsection{Attention Mechanisms}
Attention Mechanisms allow models to focus on prominent features while suppressing irrelevant information. These approaches increase performance across a wide range of applications by dynamically emphasizing crucial spatial and channel-specific information. The Convolutional Block Attention Module (CBAM) \citep{CBAM} uses a sequential method, first applying channel attention to identify relevant features, then spatial attention to determine where emphasis is needed in the feature maps. This combination is especially useful for tasks like small object detection and disease recognition \citep{aerospace11080605}. Spatial Attention Mechanisms improve CBAM by fine-tuning spatial links within feature maps, allowing models to effectively focus on regions of interest. This is especially effective in dense object detection applications, where spatial dependencies are critical \citep{CBAM}.
Both these attention mechanisms enhance the representation capabilities of lightweight networks, making them suitable for use in environments with limited computational resources. The primitive attention mechanisms such as Efficient Channel Attention (ECA) \citep{ECA} and Coordinate Attention (CA) \citep{CA} improve the balance between computational cost and performance. \\
For real-time applications, ECA is an ideal solution as it simplifies channel attention by utilizing local cross-channel interactions, which leads to lowering computing overhead while maintaining high accuracy. Coordinated Attention, on the other hand, uses coordinate encoding to merge channel and spatial attention. This method enables finer-grained feature extraction with less computing cost. The development of lightweight attention modules, such as Squeeze-and-Excitation (SE) \citep{SE}, CBAM, ECA, and CA, highlights their important role in achieving a balance between efficiency and effectiveness. These advancements are especially important in situations where resources are limited and high performance is needed alongside low computational complexity. However, existing approaches do not incorporate frequency-domain features into the attention process, which limits their ability to capture frequency-specific patterns crucial for tasks such as flood detection. By overlooking the frequency domain, current methods may miss valuable information that could enhance their performance under diverse and challenging conditions. Consequently, integrating frequency-domain features into the attention mechanism presents a promising direction for improving the robustness and performance of classification tasks.
\section{Problem Statement} \label{section3}
Let \( D = \{ (V_i, y_i) \}_{i=1}^N \) be a dataset containing \( N \) samples, where each sample consists of \( V_i \), an image corresponding to the \( i \)-th sample and \( y_i \), the label for the \( i \)-th sample. Here, \( y_i = 0 \) represents a non-flood image, and \( y_i = 1 \) represents a flood image.
 For each image \( V_i \), we generate a corresponding textual description \( T_i \) using a Vision-Language Model. 
Given an image \( V \) and its corresponding generated textual description \( T \), the objective is to classify whether the image represents a flood event or not. The task can be formally expressed as shown in \autoref{eq0}:
\begin{equation} \label{eq0}
   \hat{y_i} = \arg\max_{c \in \{0, 1\}} \, P(y_i = c \mid V_i, T_i) 
\end{equation}
where  \( \hat{y_i} \) is the predicted label for the \( i \)-th sample, \( c \in \{0, 1\} \), where \( c = 0 \) for non-flood images and \( c = 1 \) for flood-related images, \( P(y_i = c \mid V_i, T_i) \) is the conditional probability that the \( i \)-th sample belongs to class \( c \), given the image \( V_i \) and the corresponding generated textual description \( T_i \).\\\\
\textbf{\textit{{ Research Objectives}}}\\
The research objectives of the proposed approach are as follows: 
\begin{itemize}
    \item To investigate whether the Hierarchical Cross-Modal Gated Attention Mechanism effectively captures intra-modal and inter-modal interactions between image and text modalities, ensuring optimal feature fusion and improving the model’s flood prediction capabilities.
    
    \item To assess how the Frequency-Modulated Spatial Attention improves the model's ability to capture both local spatial relationships and global frequency patterns, enhancing flood detection performance.
    
    \item To investigate whether integrating frequency-domain features within a channel attention framework enhances the model's ability to focus on salient channels, improving feature representation and model performance in classification tasks.

     \item To assess how the Reverse Feature Harmonization technique refines features and integrates these refined features with visual data, optimizing the learning process and enhancing flood classification performance.
\end{itemize}

\begin{figure*}[pos=ht]
    \centering
    \includegraphics[width=\linewidth]{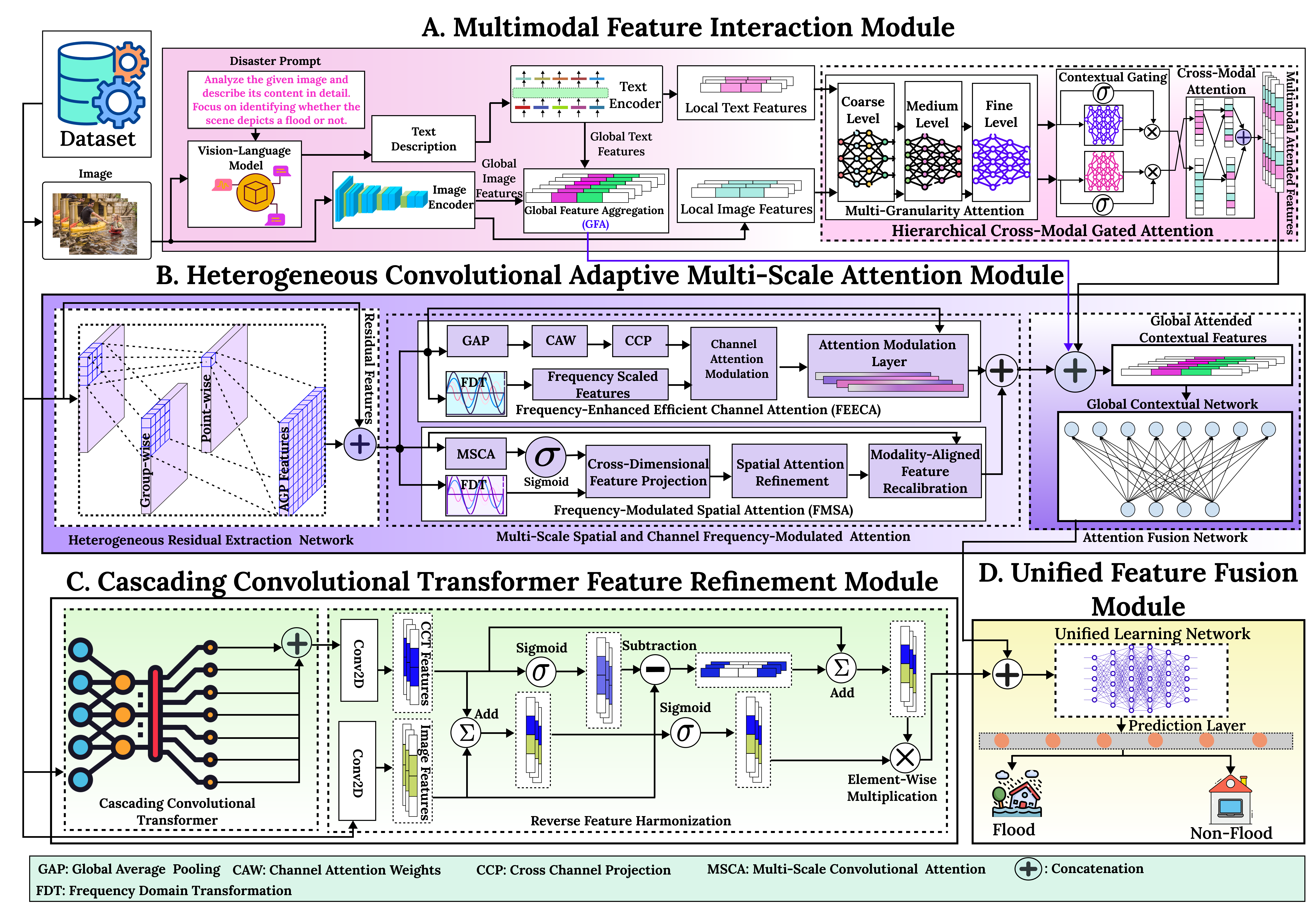}
    \caption{Illustration of XFloodNet for flood prediction, comprising four key modules: Multimodal Feature Interaction Module, Heterogeneous Convolutional Adaptive Multi-scale Attention Module (HCAMAM), Cascading Convolutional Transformer Feature Refinement Module (CCTFRM), and Unified Feature Fusion Module (UFFM). The framework utilizes hierarchical cross-modal gated attention for intra-modal and inter-modal relationships. HCAMAM applies to group and point convolutions, along with Multi-Scale Spatial and Channel Frequency-Modulated Attention for multi-scale feature extraction and enhanced feature interactions. The CCTFRM refines the features via the Reverse Feature Harmonizer, and the UFFM aggregates features from the preceding modules for final flood prediction.}
    \label{archtecture}
\end{figure*}
\section{Methodology} \label{sec3}
Our proposed framework illustrated in \cref{archtecture} is built upon four key modules: Multimodal Feature Interaction Module (MFIM), Heterogeneous Convolution Adaptive Multi-scale Attention Module (HCAMAM), Cascading Convolutional Transformer Feature Refinement Module (CCTFRM), and Unified Feature Fusion Module (UFFM). The process starts with the MFIM, which employs zero-shot prompts with a Vision-Language Model to generate text descriptions from the given image without requiring any task-specific fine-tuning. The textual and visual features undergo hierarchical cross-modal gated attention for intra-modal and inter-modal interactions between modalities. The HCAMAM employs group convolution for localized feature extraction and point convolution for pixel-level fusion. These features are passed to Multi-Scale Spatial and Channel Frequency-Modulated Attention, which operates across multiple scales and frequencies to extract fine-grained and global contextual information. The frequency features in the attention process enhance the models' focus on spatially relevant regions and channels. The CCTFRM leverages gated convolutional neural networks and transformers for class refinements. The refined features are passed to the Reverse Feature Harmonizer, which applies gated subtraction and adaptive scaling between visual and refined features for enhanced model performance. Lastly, UFFM aggregates feature vectors from MFIM, HCAMAM, and CCTFRM, producing a unified representation processed by the prediction layer to generate flood predictions.

\subsection{Multimodal Feature Interaction Module}
The Multimodal Feature Interaction Module (MFIM) systematically extracts, embeds, and fuses textual and visual information to produce a rich multimodal representation. This module is essential for creating textual descriptions using Vision-Language Models (VLMs) from visual data. The processing pipeline includes independent feature extraction of each modality, the fusion of their global representations and the extraction of local features for feature interaction through a hierarchical cross-modal gated attention mechanism. This module describes the major steps in the following manner:
\subsubsection{ Feature Extraction}
Textual descriptions produced by VLMs are generated in response to zero-shot prompts. The prompt is as: \textit{``Analyze the given image and describe its content in detail. Focus on identifying what the scene depicts in an image.''}\\
These prompts assist the model in generating textual descriptions of the visual inputs provided to it. These textual descriptions ($X_t$) are processed by a text encoder that transforms the input sequence into a feature representation. The textual features, denoted as \({H}_t \in \mathbb{R}^{n_t \times d_t} \), are obtained from the textual encoder, where $n_t$ denotes sequence length and \( d_t \) represents the dimensionality of the embedding space. These embeddings, representing token-level features, are termed local textual features, encompassing detailed information from the input sequence.  
Similarly, the input image \( X_i \) is subjected to preprocessing, typically involving normalization, resizing, and scaling to produce uniform and fixed-size inputs. The preprocessed images are subsequently encoded into a feature vector representation by an image encoder. The feature vector representation is denoted as ${H}_i \in \mathbb{R}^{H \times W \times d_i}$, where $H$ and $W$ represent the height and width of the encoded visual features and \( d_i \) represents the feature dimensionality. These features are termed local visual features, which are region-level descriptors containing detailed information about the image. The local visual features are subsequently interacted and fused with the associated local textual features in this module to enhance multimodal analysis. Moreover, we extract global features to encapsulate high-level semantic information across modalities, providing a holistic representation of their interactions. These features form the foundation for the subsequent global feature extraction process.
\subsubsection{Global Feature Extraction} \label{global}
The global feature extraction process plays a crucial role in ensuring a comprehensive understanding of the underlying semantic information from both modalities. For textual features, employing the \texttt{[CLS]} token output or applying mean-pooling across all token embeddings effectively encapsulates the semantic essence of the text. Similarly, global average pooling on visual features condenses the spatial details into a single, compact feature vector. The resulting global textual feature ${H}_G^t \in \mathbb{R}^{d_t}$ and global visual feature ${H}_G^i \in \mathbb{R}^{d_i}$, with $d_t$ and $d_i$ as their respective feature dimensions, provide a high-level representation of each modality. By concatenating these global features into a unified multimodal representation ${H}_G^{\text{concat}} \in \mathbb{R}^{d_t + d_i}$, the framework ensures that the most salient information from both modalities is retained. This unified representation serves as a high-level abstraction, providing the global context necessary for downstream tasks and enabling the model to make informed decisions. \\
The extracted local textual features ${H}_t \in \mathbb{R}^{n_t \times d_t}$ and local visual features ${H}_i \in \mathbb{R}^{H \times W \times d_i}$ serve as inputs to the Hierarchical Cross-Modal Gated Attention Mechanism. By leveraging these local features, the model can effectively capture hidden cross-modal dependencies, enabling a deeper understanding of the interactions between textual and visual data. This process enhances the model's ability to reason about complex multimodal information, paving the way for improved performance in tasks requiring fine-grained feature alignment and integration.
\subsubsection{Hierarchical Cross-Modal Gated Attention Mechanism}
\begin{figure*}[pos=ht]
    \centering
    \includegraphics[width=\linewidth]{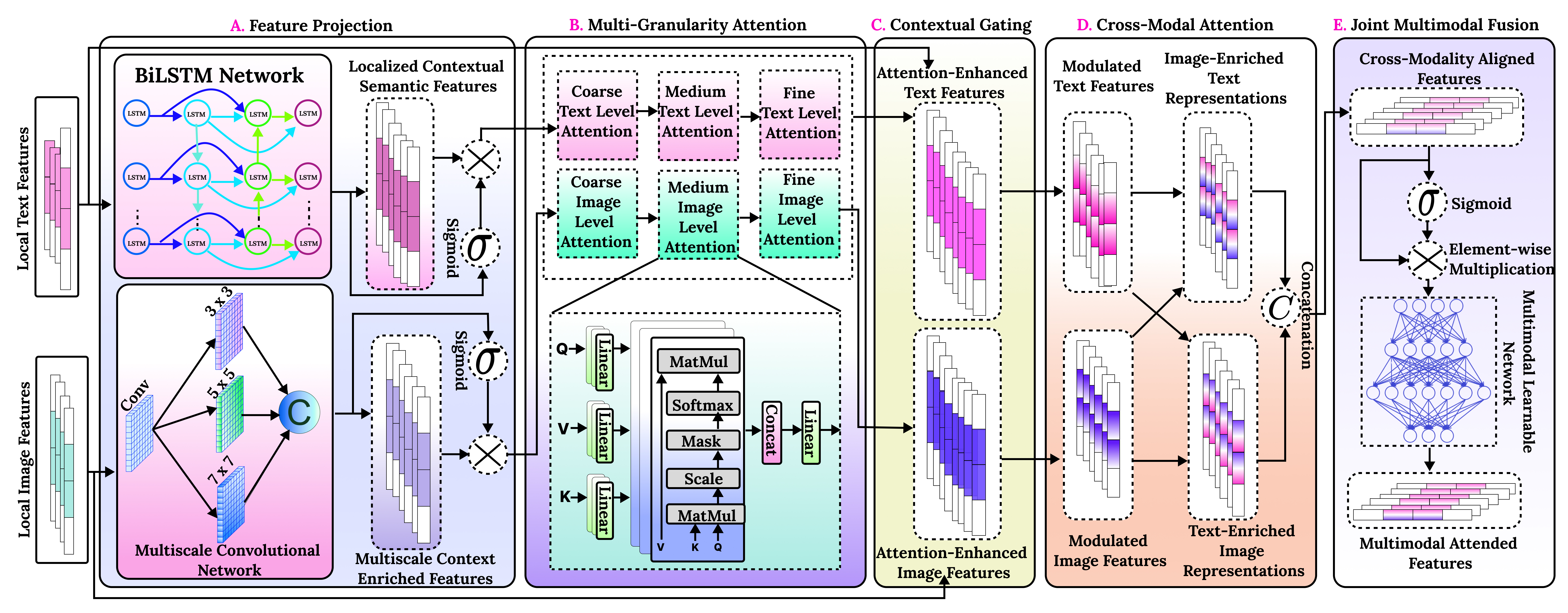}
    \caption{Overview of the Hierarchical Cross-Modal Gated Attention Mechanism, a framework for fusing textual and visual features through multi-level attention. The mechanism consists of five key steps: (a) Feature Projection, ensuring dimensional compatibility between modalities; (b) Multi-Granularity Attention Representation, which refines intra-modal dependencies across coarse, medium, and fine levels; (c) Contextual Gating, which filters and modulates information flow to highlight salient features; (d) Cross-Modal Attention, enabling bidirectional attention to capture complex relationships between text and image; and (e) Joint Multimodal Fusion, which aggregates cross-modal attention into a unified representation.}
    \label{attention}
\end{figure*}

The Hierarchical Cross-Modal Gated Attention Mechanism (HCGAM), as illustrated in \autoref{attention}, is a structured attention framework that processes and fuses features from multiple modalities, such as text and image. This mechanism is designed to facilitate fine-grained interactions between the textual and visual modalities by dynamically modulating attention weights across multiple levels of abstraction. The multi-level gated attention mechanism selectively attends to salient regions in both the text and image, ensuring that complementary information from both modalities is effectively aligned and integrated.\\
HCGAM consists of these steps: (a) {Feature Projection}: Project textual and visual modalities into a unified representational space, ensuring dimensional compatibility; (b) {Multi-Granularity Attention:} Decompose attention into multiple levels such as Coarse, Medium, and Fine allowing progressive refinement of intra-modal dependencies; (c) {Contextual Gating:} Introduce a non-linear gating mechanism that modulates and filters information flows, highlighting salient features; (d) {Cross-Modal Attention:} Facilitate bidirectional attention flow between text and image to capture complex inter-modal relationships; (e) { Joint Multimodal Fusion:} Concatenate cross-modal attention to form a unified feature representation for multimodal tasks.
\paragraph{Feature Projection:}
The textual and visual features are represented as \({H}_t \in \mathbb{R}^{n_t \times d_t}\) and \({H}_i \in \mathbb{R}^{H \times W \times d_i}\), respectively. To effectively align and integrate information across modalities, it is essential to project these features into a shared feature space $d_\text{se}$. This projection ensures that both text features \({H}_t \in \mathbb{R}^{n_t \times d_{\text{se}}} \) and image representations \({H}_i \in \mathbb{R}^{H \times W \times d_\text{se}} \) are compatible for further modal interactions, allowing the model to effectively capture dependencies between the modalities.\\
The local textual features ${H}_t \in \mathbb{R}^{n_t \times d_\text{se}}$ are further processed through a Bidirectional Long Short-Term Memory (BiLSTM) network. This step captures contextual dependencies in both forward and backward directions, enriching the representation of the textual features.  The BiLSTM output, referred to as Localized Contextual Semantic Features, better encodes sequential information, making it more suitable for multimodal interactions in downstream tasks. The localized image features ${H}_i \in \mathbb{R}^{H \times W \times d_\text{se}}$ undergo multi-scale operations using convolutional kernels of sizes $3 \times 3$, $5 \times 5$, and $7 \times 7$, which enable the model to capture visual patterns at various granularities. The multi-scale features are concatenated to form a representation known as Multiscale Context Enriched Features, enhancing the model's ability to reason over complex visual data.\\
The spatial dimensions \( H \times W \) of the image features are flattened into a single dimension \( n_i = H \times W \), representing the total number of spatial regions in the image. The resulting image features are then denoted as \( {H}_i \in \mathbb{R}^{n_i \times d_\text{se}} \), where \( n_i \) corresponds to the number of spatial regions, and \( d_\text{se} \) is the dimensionality of the shared feature space. This transformation allows for a uniform representation, making the visual features compatible for further processing within the model’s multimodal architecture. The \({H}_t\) and \({H}_i\) are processed through element-wise multiplication with their respective sigmoid-activated counterparts as shown in \autoref{eq1a} and \autoref{eq1b}. The resultant feature vectors \({H}_t' \in \mathbb{R}^{n_t \times d_\text{se}} \) and \({H}_i' \in \mathbb{R}^{n_i \times d_\text{se}}\) are represented as:
\begin{subequations}
\begin{align}
    {H}_t' &= \left( {H}_t \odot \sigma \left( {H}_t \cdot {W}_t + {b}_t \right) \right) \label{eq1a} \\
    {H}_i' &= \left( {H}_i \odot \sigma \left( {H}_i \cdot {W}_i + {b}_i \right) \right) \label{eq1b}
\end{align}
\end{subequations}
where \( \sigma \) denotes the sigmoid function, \( \odot \) denotes element-wise multiplication and $W_t, W_i$ are all trainable weighted matrices and $b_t, b_i$ are biases. This process captures complex interactions between the input features, refining the output through a combination of transformations. These features are further refined via a multi-granularity attention mechanism.

\paragraph{Multi-Granularity Attention Mechanism:}
The multi-granularity attention mechanism is designed to capture and enhance intra-modality relationships within the textual and visual modalities through three distinct stages: coarse-level, medium-level, and fine-level attention. Each stage employs multi-headed attention to facilitate the learning of complex dependencies and interactions among the features of each modality, thereby improving the model's ability to represent the complex details and patterns within the data.

\subparagraph{\textit{Coarse-level Attention:}}
At this initial stage, the attention mechanism focuses on establishing a high-level understanding of the input features by employing a reduced number of attention heads. This phase captures broad patterns within the modality, allowing for efficient processing of large-scale features. The coarse-level attention sets the foundation for subsequent stages.
For Coarse-level Attention, the number of attention heads is \( \frac{h}{2} \), where \( h \) is the total number of attention heads. The dimensionality of each attention head, denoted as \( d_\text{head} \). The relationship is expressed in \autoref{eq2}:

\begin{equation} \label{eq2}
    d_\text{head} = \frac{d_\text{se}}{\frac{h}{2}} = \frac{2 \cdot d_\text{se}}{h}
\end{equation}
This ensures that the total feature dimensionality \( d_\text{se} \) is consistently distributed across the \( \frac{h}{2} \) attention heads in the Coarse-level Attention mechanism.
Let \( W_Q^{\text{c}}, W_K^{\text{c}}, W_V^{\text{c}} \in \mathbb{R}^{d_{\text{se}} \times {d_{\text{head}}}}\) represent the trainable projection matrices for the query, key, and value, respectively. For the text features \( H_t \in \mathbb{R}^{n_t \times d_{\text{se}}} \), the coarse-level attention mechanism computes the query, key, and value as: ($Q_t^{\text{c}} = H_t W_Q^{\text{c}}$, $K_t^{\text{c}} = H_t W_K^{\text{c}}$, and $ V_t^{\text{c}} = H_t W_V^{\text{c}}) \in \mathbb{R}^{n_t \times d_{\text{head}}}$,  respectively. The visual counterparts share the same dimensions as the textual modality. The coarse-level attention for each attention head for text and image modality is computed as depicted in \autoref{eq3a} and \autoref{eq3b}, respectively.
\begin{subequations}
\begin{align}
    A_\text{t}^{\text{c}} &= \text{softmax} \left( \frac{Q_t^{\text{c}} ({K_t^{\text{c}}})^T}{\sqrt{\frac{2d_{\text{se}}}{h}}} \right) V_t^{\text{c}} \in \mathbb{R}^{n_t \times d_{\text{head}}} \label{eq3a} \\
    A_\text{i}^{\text{c}} &= \text{softmax} \left( \frac{Q_i^{\text{c}} ({K_i^{\text{c}}})^T}{\sqrt{\frac{2d_{\text{se}}}{h}}} \right) V_i^{\text{c}} \in \mathbb{R}^{n_i \times d_{\text{head}}} \label{eq3b}
\end{align}
\end{subequations}
Once all \( \frac{h}{2} \) attention heads have computed their respective outputs, these are concatenated along the feature dimensions. The concatenated output is then passed through a linear transformation (weight matrix \( W^O \in \mathbb{R}^{d_{\text{se}} \times d_{\text{se}}}  \) ) to map it back to the original feature space as depicted in \autoref{eq4a} and \autoref{eq4b}.
\begin{subequations}
\begin{align}
Att_\text{t}^{\text{c}} = \text{concat}(A_\text{t,1}^{\text{c}}, A_\text{t,2}^{\text{c}}, \dots, \text{A}_{t,\frac{h}{2}}) \cdot W^O \in \mathbb{R}^{n_t \times d_{\text{se}}} \label{eq4a} \\
Att_\text{i}^{\text{c}} = \text{concat}(A_\text{i,1}^{\text{c}}, A_\text{i,2}^{\text{c}}, \dots, \text{A}_{i,\frac{h}{2}}) \cdot W^O \in \mathbb{R}^{n_i \times d_{\text{se}}} \label{eq4b}
\end{align}
\end{subequations}
Building on the broad patterns captured by Coarse-level Attention, these features are subsequently refined and enriched through Medium-level Attention, which delves deeper into intermediate-level dependencies to enhance feature representation.
\subparagraph{\textit{ Medium-level Attention:}} The medium-level attention refines the representations by focusing on more detailed relationships within the modalities. This stage enhances the representation of features obtained from the coarse level, allowing for a deeper understanding of the internal structures and dependencies present within the textual or visual input. At the medium level, the number of heads is \( h \). The relationship between $d_\text{head}$ and $d_\text{se}$ at medium-level attention can be calculated as \(d_\text{head} = \frac{d_\text{se}}{h}\). The input to the medium-level attention is the coarse-level attended output.  For the text features, the trainable weight matrices \( W_Q^{\text{m}} \), \( W_K^{\text{m}} \), and \( W_V^{\text{m}} \), each of size \( \mathbb{R}^{{d_{\text{se}}} \times d_{\text{head}}} \), used to compute the corresponding queries, keys, and values as $(Q_t^{\text{m}} = A_\text{tt}^{\text{c}} W_Q^{\text{m}}$, $ K_t^{\text{m}} = A_\text{tt}^{\text{c}} W_K^{\text{m}}$, and $V_t^{\text{m}} = A_\text{tt}^{\text{c}} W_V^{\text{m}}) \in \mathbb{R}^{n_t \times d_{\text{head}}}$, respectively. The visual features share identical dimensions with the textual modality. The medium-level attention for each head, across both text and image modalities, is computed as illustrated in \autoref{eq5a} and \autoref{eq5b}, respectively:
\begin{subequations}
\begin{align}
    A_t^{\text{m}} = \text{softmax} \left( \frac{Q_t^{\text{m}} ({K_t^{\text{m}}})^T}{{\sqrt{\frac{d_{\text{se}}}{h}}}} \right) V_t^{\text{m}} \in \mathbb{R}^{n_t \times {d_{\text{head}}}} \label{eq5a}
\\
    A_i^{\text{m}} = \text{softmax} \left( \frac{Q_i^{\text{m}} ({K_i^{\text{m}}})^T}{{\sqrt{\frac{d_{\text{se}}}{h}}}} \right) V_i^{\text{m}} \in \mathbb{R}^{n_i \times {d_{\text{head}}}} \label{eq5b}
    \end{align}
\end{subequations}
where \(A_t^{\text{m}}, A_i^{\text{m}}  \) represents medium-level attended textual and visual features of each head, respectively. After concatenating all $h$ heads and putting all the values in the \autoref{eq4a} and  \autoref{eq4b}, the final medium-level textual and visual attended features are denoted as $Att_\text{t}^{\text{m}}  \in \mathbb{R}^{n_t \times {d_{\text{se}}}}$ and $Att_\text{i}^{\text{m}}  \in \mathbb{R}^{n_i \times {d_{\text{se}}}}$, respectively. At the medium level, the model is allowed to capture more complex and detailed relationships within the input, thereby improving its ability to interpret both text and image modalities with greater accuracy.

\subparagraph{\textit{Fine-level Attention:}} At the fine level, the model enhances its ability to identify detailed patterns and intricate relationships within the multimodal inputs by doubling the number of attention heads to \( 2h \). The relationship between $d_\text{se}$ and $d_\text{head}$ can be calculated as \(d_\text{head}=\frac{d_\text{se}}{2h}\). This stage refines the representations derived from the medium-level attention mechanisms, enabling the model to capture precise details and complex interactions essential for accurate feature representation.
Let the query, key, and trainable weight matrices for fine-level attention be 
\( W_Q^{\text{f}}, W_K^{\text{f}}, W_V^{\text{f}} \in \mathbb{R}^{d_\text{se} \times d_\text{head}} \), respectively. The query, key, and value projections for fine-level textual features are computed as: 
\((Q_t^{\text{f}} = A_t^{\text{m}} W_Q^{\text{f}}, K_t^{\text{f}} = A_t^{\text{m}} W_K^{\text{f}},  V_t^{\text{f}} = A_t^{\text{m}} W_V^{\text{f}}) \in \mathbb{R}^{n_t \times d_\text{head}}\). The visual features have dimensions identical to those of the textual modality. The fine-level attended textual and visual features for each attention head are computed as described in \autoref{eq6a} and \autoref{eq6b}, respectively:

\begin{subequations} 
\begin{align}
A_t^{\text{f}} = \text{softmax}\left( \frac{Q_t^{\text{f}} ({K_t^{\text{f}}})^T}{\sqrt{\frac{d_{\text{se}}}{2h}}} \right) V_t^{\text{f}} \in \mathbb{R}^{n_t \times {d_{\text{head}}}} \label{eq6a}
\\
A_i^{\text{f}} = \text{softmax}\left( \frac{Q_i^{\text{f}} ({K_i^{\text{f}}})^T}{\sqrt{\frac{d_{\text{se}}}{2h}}} \right) V_i^{\text{f}} \in \mathbb{R}^{n_i \times {d_{\text{head}}}} \label{eq6b}
\end{align}
\end{subequations}
where \( A_t^{\text{f}}\) and \( A_i^{\text{f}} \) denote the fine-level attended textual and visual features for each attention head, respectively. After concatenating the features from all \(2h\) heads and applying the transformation in \autoref{eq4a} and  \autoref{eq4b}, the final fine-level attended features for text and image are represented as \( Att_\text{t}^{\text{f}} \in \mathbb{R}^{n_t \times d_{\text{se}}} \) and \( Att_\text{i}^{\text{f}} \in \mathbb{R}^{n_i \times d_{\text{se}}} \), respectively. 

This fine-level attention mechanism enhances the feature representations of both text and image modalities by capturing intricate patterns and relationships, thereby improving the model's ability to understand and process the input modalities with greater precision and comprehensiveness.

\paragraph{Contextual Gating}:
The attention-enhanced textual features \( Att_\text{t}^{\text{f}} \in \mathbb{R}^{n_t \times d_{\text{se}}} \) and visual features \( Att_\text{i}^{\text{f}} \in \mathbb{R}^{n_i \times d_{\text{se}}} \) are processed through a contextual adaptive gating mechanism. This mechanism applies learned gating parameters to selectively enhance the features, considering both global and local contextual information. The detailed gating operation is shown in \autoref{eq7a} and \autoref{eq7b}:

\begin{subequations} 
\begin{align}
    G(A_t) &= \sigma \left( \mathcal{L}\left( H_t W_t + b_t \right) \right) \odot \left( Att_t^{\text{f}} \right)  \label{eq7a}\\
    G(A_i) &= \sigma \left( \mathcal{L}\left( H_i W_i + b_i \right) \right) \odot \left( Att_i^{\text{f}} \right) \label{eq7b}
\end{align}
\end{subequations}
where \( G(A_t) \in \mathbb{R}^{n_t \times d_{\text{se}}} \) and \( G(A_i) \in \mathbb{R}^{n_i \times d_{\text{se}}} \) represent the modulated textual and visual features, respectively, \( \mathcal{L}\) denotes layer normalization. Here, \( H_t \in \mathbb{R}^{n_t \times d_\text{se}} \) and \( H_i \in \mathbb{R}^{n_i \times d_\text{se}} \) are the textual and visual features in the same embedding space. \( W_t \in \mathbb{R}^{n_\text{se} \times n_t} \) and \( W_i \in \mathbb{R}^{d_\text{se} \times n_i} \) are learnable weight matrices, and \( b_t \) and \( b_i \) are the corresponding bias terms. 
This adaptive gating mechanism allows the model to dynamically adjust the strength of feature interactions, refining both textual and visual representations by focusing on relevant patterns while suppressing noise. The resulting gated features, which are refined for each modality, are now in an optimal form for subsequent cross-modality attention. This mechanism facilitates the alignment of textual and visual information by capturing intricate relationships and correlations between the two modalities. By leveraging the refined features, the cross-modality attention process improves the model's ability to fuse complementary information from both domains, leading to a more cohesive and enriched representation that enhances downstream task performance.
\paragraph{Cross-Modal Attention:}

The modulated textual features and modulated image features are processed through a cross-modality attention mechanism, enabling interactions between the text and image modalities. This mechanism allows each modality to refine its representation by attending to the complementary modality.
For text-to-image attention, the query \( Q_t \in \mathbb{R}^{{n_t} \times d_{\text{se}}} \) is derived from the textual features \( G(A_t) \), while the key \( K_i \in \mathbb{R}^{{n_i} \times d_{\text{se}}} \) and value \( V_i \in \mathbb{R}^{{n_i} \times d_{\text{se}}} \) are derived from the image features \( G(A_i)\). Similarly, for image-to-text attention, the query \( Q_i \in \mathbb{R}^{{n_i} \times d_{\text{se}}} \) is derived from \(G(A_i)\), while the key \( K_t \in \mathbb{R}^{{n_t} \times d_{\text{se}}}  \) and value \( V_t \in \mathbb{R}^{{n_t} \times d_{\text{se}}} \) are derived from \( G(A_t) \). These transformations for text modality are defined as \(Q_t = G(A_t)W_Q^t,\quad K_t = G(A_t) W_K^t,\quad V_t = G(A_t)W_V^t\). Similarly, these transformations for image modality are defined as \( Q_i = G(A_i) W_Q^i,\quad  K_i =  G(A_i) W_K^i, \quad V_i =  G(A_i)W_V^i \). The trainable learnable weight matrices are defined as, \( W_Q^t, W_K^t, W_V^t \in \mathbb{R}^{{n_t} \times d_{\text{se}}}\) and \( W_Q^i, W_K^i, W_V^i \in \mathbb{R}^{{n_i} \times d_{\text{se}}}\).

The text-to-image (\(\text{Att}_{t \rightarrow i} \in \mathbb{R}^{n_t \times d_{\text{se}}}\)) and image-to-text attention \((\text{Att}_{i \rightarrow t} \in \mathbb{R}^{n_i \times d_{\text{se}}})\) outputs are computed as shown in \autoref{eq9a} and \autoref{eq9b}, respectively:

\begin{subequations} 
\begin{align}
   \text{Att}_{t \rightarrow i} = \text{Softmax}\left( \frac{Q_t K_i^\top}{\sqrt{d_k}} \right) V_i  \in \mathbb{R}^{n_t \times d_{\text{se}}} \label{eq9a}\\
\text{Att}_{i \rightarrow t} = \text{Softmax}\left( \frac{Q_i K_t^\top}{\sqrt{d_k}} \right) V_t  \in \mathbb{R}^{n_i \times d_{\text{se}}} \label{eq9b}
 \end{align}
\end{subequations}

where \( d_k \) is the scaling factor (dimension of the key vectors) used to normalize the attention scores. The cross-modal attention mechanism enables detailed and complementary feature interactions, enhancing the joint representation of the input modalities.
\paragraph{Joint Multimodal Fusion:}
These features are subsequently concatenated to form a unified representation, $A_\text{concat} \in \mathbb{R}^{(n_t+n_i) \times d_{\text{se}}}$. To introduce non-linearity and enhance feature expressivity, a sigmoid activation function is applied to the concatenated cross-modal features. Following the activation, element-wise multiplication is performed between the activated features and the original cross-modal features to selectively emphasize significant interactions. This process amplifies the most relevant information while suppressing less important interactions, yielding refined cross-modal representations.  \\
The refined cross-modal representations, $A_\text{concat}$, are processed through a Multimodal Learnable Network (MLN), an advanced framework tailored to enhance the joint feature space by capturing both modality-specific characteristics and shared interdependencies. The MLN employs trainable parameters to adaptively process and integrate the heterogeneous features, thereby ensuring comprehensive representation learning across modalities.\\
The resulting Multimodal Attended Features effectively encapsulate intricate, modality-aware interactions, preserving complementary and discriminative information critical for downstream tasks. These learned features serve as robust inputs for subsequent applications, including classification or information retrieval, facilitating high-level performance by leveraging the enriched cross-modal representations.
\subsection{Heterogeneous Convolutional Adaptive Multi-Scale Attention Module}
The Heterogeneous Convolutional Adaptive Multi-Scale Attention Module (HCAMAM) is a pivotal component designed to enhance the model's ability to extract and fuse features across spatial and channel dimensions using frequency components. HCAMAM integrates advanced mechanisms, including Heterogeneous Residual Extraction Network (HREN), Multi-Scale Spatial and Channel Frequency-Modulated Attention (MSCFMA), and an Attention Fusion Block, to deliver a robust feature representation for complex visual tasks such as flood classification.\\
HREN captures point-wise and group-wise pixel information, effectively extracting coarse-to-fine-grained details from visual data. Combining these granular details ensures the model can process both localized and high-level contextual information. MSCFMA is composed of two specialized components:
Frequency-Enhanced Efficient Channel Attention (FEECA) emphasizes channel-wise information by leveraging frequency-based enhancements to boost feature extraction in critical regions. Frequency-modulated spatial Attention (FMSA) complements FEECA by focusing on spatial information and modulating it through frequency-driven refinements. The outputs from FEECA and FMSA are concatenated, forming a unified feature representation that seamlessly integrates spatial and channel dimensions for improved multimodal understanding.\\
To ensure holistic feature integration, Attention Fusion Block concatenates features from MSCFMA with MFIM. This fusion bridges cross-modal multimodal relationships, as well as local and global contexts, leading to seamless integration of multimodal and multi-scale information into the model architecture.
By leveraging these interdependent components, HCAMAM unifies representations, empowering the model to achieve precise and reliable outcomes in flood classification scenarios. Its ability to operate across multiple scales and dimensions ensures that the model captures both fine-grained details and broader contextual relationships, making it particularly effective for complex disaster-related visual tasks.

\subsubsection{Heterogeneous Residual Extraction Network}

The Heterogeneous Residual Extraction Network (HREN) is designed to enhance feature extraction through a combination of residual connections and diverse convolutional operations. This architecture incorporates both groupwise and pointwise convolutions, significantly improving the network's learning capacity while maintaining effective gradient flow during training. This combination also promotes efficient parameter utilization and robustness against overfitting.\\
Let \(H\) represent the height, \(W\) the width, \(C_{\text{in}}\) the number of input channels, \(C_{\text{out}}\) the number of output channels, \(G\) the number of groups for groupwise convolution, and \(K\) the kernel size. Initially, a groupwise convolution is applied to the input image \(X_i \in \mathbb{R}^{H \times W \times C_{\text{in}}}\), yielding an output feature map \(X_1 \in \mathbb{R}^{H \times W \times C_{\text{out}}}\). Subsequently, a pointwise convolution is performed, producing another feature map \(X_2 \in \mathbb{R}^{H \times W \times C_{\text{out}}}\). The final output denoted as ${X_f}'$ represents the aggregate point-group-wise  (AGP) Features of the HREN, which is computed as shown in  \autoref{eq10}:
\begin{equation} \label{eq10}
X_f' = \underbrace{\sum_{g=1}^{G} W_g \ast X_1}_{\text{Groupwise Convolution}}
+ \underbrace{W_p \ast \big(\text{BN}(X_2)\big)}_{\text{Pointwise Convolution}}
\end{equation}
where \(W_g \in \mathbb{R}^{K \times K \times \frac{C_{\text{in}}}{G} \times \frac{C_{\text{out}}}{G}}\) are the groupwise convolution kernels, \(W_p \in \mathbb{R}^{1 \times 1 \times C_{\text{out}} \times C_{\text{out}}}\) is the pointwise convolution kernel,
\(\text{BN}(X)\) represents batch normalization applied to \(X_2\), \( K \times K \) refers to the size of the kernel used in the convolution operation. The dimensions of \( W_g \) also account for the number of input channels (\( C_{\text{in}} \)) and the number of output channels (\( C_{\text{out}} \)). Since groupwise convolution divides the input into \( G \) groups, \( \frac{C_{\text{in}}}{G} \) and \( \frac{C_{\text{out}}}{G} \) are the numbers of input and output channels per group, respectively. A residual connection is added to preserve low-level features, improve gradient flow, and enhance feature refinement.
\begin{equation} \label{eq11}
    X_f = X_f' + \text{Res}(X_i) \in \mathbb{R}^{H \times W \times C_{\text{out}}}
\end{equation}
In \autoref{eq11}, $X_f$ refers to the final extracted features from the HREN module. \(\text{Res}(X_i)\) denotes the residual connection.
The HREN offers several advantages. It enhances feature learning through the integration of diverse convolution techniques, allowing the network to capture a wider range of features from input data. This flexibility improves the network's ability to learn complex patterns, thereby enhancing performance in classification tasks. In brief, the output features extracted from the HREN module retain the spatial and channel dimensions of the input image while benefiting from the enhanced feature extraction process that combines groupwise, pointwise convolutions, and residuals. \\
In brief, the output features extracted from the Heterogeneous Residual Extraction Network (HREN) module retain the spatial and channel dimensions of the input tensor while benefiting from the enhanced feature extraction process that combines groupwise and pointwise convolutions, along with residual connections. This ensures that both local and global dependencies are captured effectively. Subsequently, these extracted features undergo a frequency-enhanced channel attention mechanism. The goal of this next stage is to refine the feature maps capable of focusing on the most informative regions and channels, allowing for better feature representation and more effective learning. 

\subsubsection{Multi-Scale Spatial and Channel Frequency Modulated Attention}
We introduce a Multi-Scale Spatial and Channel Frequency-Modulated Attention (MSCFMA) mechanism that operates across multiple scales, enabling the model to extract both fine-grained and global contextual information from flood images. It integrates spatial attention to identify key regions of interest and channel attention to modulate features across different frequency bands, facilitating the detection of flood indicators at varying spatial resolutions and channel frequencies.
The incorporation of frequency-domain features in both the spatial and channel dimensions allows for the explicit modeling of frequency-specific patterns. By attending to these frequency-modulated features, the model is able to effectively distinguish flood indicators from background noise, improving its robustness and accuracy in detecting floods under diverse and challenging conditions. The two submodules of MSCFMA are Frequency-Enhanced Efficient Channel Attention and Frequency-Modulated Spatial Attention.
\paragraph{\textbf{Frequency-Enhanced Efficient Channel Attention}:}
The Frequency-Enhanced Efficient Channel Attention (FEECA) mechanism is inspired by the principles behind HiLo attention \citep{hilo}, which distinguishes high-frequency and low-frequency components to enhance feature representation in vision transformers. While HiLo improves attention efficiency by untangling fine-grained (high-frequency) and global (low-frequency) features, FEECA utilizes the 2D Fast Fourier Transform (2DFFT) to extract frequency-domain characteristics that emphasize both fine details and global structures within the input features. Unlike HiLo Attention, which separates high and low-frequency features into different attention heads, FEECA integrates these frequency-domain features within a single-channel attention mechanism. By leveraging both frequency components and channel dependencies, FEECA improves the model's ability to focus on the most salient channels while maintaining computational efficiency and a lightweight architecture.
The FEECA process can be described as follows:
\\
\subparagraph{\textit{Global Average Pooling (GAP):}}
Let the input feature map be denoted as \( X_f \in \mathbb{R}^{H \times W \times C_\text{out}} \), where \( H \) is the height, \( W \) is the width, and \( C_\text{out} \) is the number of output channels. The first operation applied to the input is global average pooling, which compresses the spatial dimensions while retaining channel-wise information as shown in \autoref{GAP}:
\begin{equation}\label{GAP}
Y_{\text{GAP}}(c) = \frac{1}{H \times W} \sum_{i=1}^{H} \sum_{j=1}^{W} X_f(i, j, c)
\end{equation}  
where \( Y_{\text{GAP}} \in \mathbb{R}^{1 \times C} \) is the pooled vector, and \( c \) indexes the channel dimension. This operation effectively reduces the spatial dimensions \((H, W)\) to a single scalar value per channel, forming a compact global descriptor. 
For simplicity, now onwards, in the rest of the manuscript, we denote \( C_\text{out} \) as \( C \).
\\
\subparagraph{\textit{Channel Attention Weights:}}
The globally pooled feature map is reshaped to \((C, 1)\), preparing it for 1D convolution along the channel dimension. A 1D convolution with a fixed kernel is then applied, producing channel attention weights. The resulting tensor has the shape \((C, 1)\), where each element reflects the relative importance of each channel. This operation enables local channel interactions, allowing the model to learn complex inter-channel relationships and dependencies from the global context. 
\\
\subparagraph{\textit{Cross-Channel Projection:}}
After obtaining the channel attention weights, we apply a fully connected layer. This step ensures that the output shape matches the input feature map, allowing for element-wise multiplication later in the process. It effectively integrates the learned attention weights into the model’s overall architecture. This cross-channel projection transforms the attention weights back to the original number of channels. We apply a cross-channel projection as shown in \autoref{eq13}:
\begin{equation} \label{eq13}
\begin{aligned}
Y_{\text{Proj}}[n, m] = & \sum_{c=1}^{C_{\text{out}}} \Bigg( 
    \sum_{i=1}^{L} \sum_{j=1}^{C_{\text{in}}} 
    Y_{\text{GAP}}[n + i - 1, j] \cdot W_{\text{conv}}[i, j, c] \\
    & + b_{\text{conv}}[c] \Bigg) \cdot W_{\text{proj}}[m, c] + b[m]
\end{aligned}
\end{equation}
where \( Y_{\text{Proj}}[n, m] \in \mathbb{R}^{C \times 1} \) represent the projected output for sample \( n \) and feature \( m \), while \( Y_{\text{GAP}}[n, j] \) denotes the GAP output, corresponding to sample \( n \) and channel \( j \). The convolutional kernel weights are denoted as \( W_{\text{conv}}[i, j, c] \), and the convolutional bias is represented by \( b_{\text{conv}}[c] \). The projection weights are \( W_{\text{proj}}[m, c] \), and the projection bias is \( b[m] \). Here, \( L \) defines the number of spatial indices \( (i) \) considered during the convolution, which helps in extracting spatial features from the input. The output feature map \( Y_{\text{Proj}} \) is then reshaped to match the desired dimensions for subsequent processing.
\\
\subparagraph{\textit{Frequency Domain Transformation:}} A unique aspect of the FEECA is the incorporation of frequency domain analysis. We apply the 2D Fast Fourier Transform (2DFFT) to the input feature map $X_f$, converting the spatial features into their frequency components. The transformation outputs capture the magnitude of the frequencies, providing a new representation that captures global and small details in the input feature map. Frequency features can enhance the model's sensitivity to patterns and textures that spatial features might miss. They improve robustness against noise and small perturbations, making the model more reliable in diverse conditions.
Next, we apply the 2D Fast Fourier Transform (2DFFT) to the input feature map \( X_f \) along the spatial dimensions for each channel as shown in \autoref{eq14}.
\begin{equation} \label{eq14}
X_{\text{freq}} = \left\lvert \text{2DFFT} \left( X_f \right) \right\rvert = \sqrt{\text{Re}^2(X_f) + \text{Im}^2(X_f)}
\end{equation}
Here, \( X_{\text{freq}} \in \mathbb{R}^{H \times W \times C} \) captures the magnitude of frequency components for all channels. \(\text{Re}(X_f)\) and \(\text{Im}(X_f)\)
represents the real and imaginary parts of the frequency components of \(X_f\).
\\
\subparagraph{\textit{Scaling Frequency Features:}} We introduce learnable frequency scaling factors, which are applied to the frequency features. The scaling factors allow the model to adaptively emphasize certain frequency components while suppressing less informative ones. This scaling enables the attention process to learn which frequency features are most relevant to the task at hand, enhancing the overall representational power of the model. We apply learned scaling factors ${S} \in \mathbb{R}^{1 \times 1 \times C}$ to dynamically emphasize certain frequency components. It is done by multiplying the scaling frequency features with the $X_\text{freq}$. The output of this process is denoted as \(SFF_{\text{scaled}} = S \odot X_{\text{freq}} \in \mathbb{R}^{H \times W \times C}\) is the frequency-modulated feature map.
\\
\subparagraph{Channel Attention Modulation:} In this phase, the output of the cross-channel attention, \( Y_{\text{proj}} \), is first multiplied with the frequency-modulated feature map, \( SFF_{\text{scaled}} \), to adaptively emphasize or suppress frequency components based on the learned attention. This is followed by the normalization of the resulting product via a sigmoid activation function, which maps the values to the range \( [0, 1] \), enabling dynamic modulation of the feature map. The formulation of this process is shown in \autoref{eq15}:
\begin{equation} \label{eq15}
Y_{\text{att}} = \sigma  \left( \sum_{c=1}^{C} Y_{\text{proj}}^{(c)} \cdot SFF_{\text{scaled}}^{(c)} \right)
\end{equation}
where \( Y_{\text{proj}}^{(c)} \) and \( SFF_{\text{scaled}}^{(c)} \) represent the feature values for the \( c \)-th channel of the respective feature maps. This process ensures that the frequency-scaled features are adaptively modulated by the cross-channel attention map, which is crucial for enhancing the model's representational capacity.
\\
\subparagraph{\textit{Attention Modulation Layer:}}
The final step in the feature refinement process involves modulating the input feature map \( X_f \) with a learned attention map \( Y_{\text{att}} \) that captures both spatial and frequency-based importance. The input features are enhanced by element-wise multiplication with the attention map, which selectively emphasizes relevant features while suppressing irrelevant ones. This can be expressed as:
\begin{equation} \label{eq16}
Y_{\text{mca}} = \mathcal{L}\left( Y_{\text{att}} \right) \odot \mathcal{L}\left( X_f \right)
\end{equation}
where \(\mathcal{L}(\cdot)\) represents layer normalization applied separately to both maps. \( Y_{\text{mca}} \in \mathbb{R}^{H \times W \times C} \) is the output feature map after attention modulation, which selectively enhances the important features based on the attention mechanism.
The FEECA seamlessly integrates channel and frequency domain analyses to refine feature representations. By leveraging frequency components alongside channel features, this approach enhances the model's ability to capture both fine-grained local structures and long-range dependencies within the data. The incorporation of frequency-domain information significantly improves the model’s sensitivity to subtle patterns while simultaneously increasing robustness to noise and variations that often degrade performance in traditional architectures.
\paragraph{\textbf{Frequency-Modulated Spatial Attention:}}
The Frequency-Modulated Spatial Attention (FMSA) module is designed to capture local spatial relationships and global frequency domain features. This is achieved by integrating multi-scale convolutional attention with frequency domain representations, enhancing the model's capacity to selectively focus on spatially relevant regions. The core components and functionalities of the FMSA module are as follows:
First, FMSA employs multi-scale convolutional filters to extract spatial features at varying granularities, effectively capturing both fine and coarse spatial relationships. To complement this, the module incorporates Frequency Domain Representation using the 2D Fast Fourier Transform (2DFFT), enabling the capture of global structural patterns and periodic information beyond the spatial domain. These spatial and frequency features are aligned and fused through a Cross-Dimensional Feature Projection, which leverages a learnable and normalization mechanism to ensure their seamless interaction. Finally, a Squeeze-and-Excitation Refinement Block is applied to recalibrate the feature maps, emphasizing the most informative features while suppressing irrelevant ones, thereby refining the attention maps and enhancing the overall feature learning process. \\
\subparagraph{\textit{Multi-Scale Convolutional Attention:}}  
The multi-scale convolutional attention mechanism applies 2D convolutions with multiple kernel sizes \( k_i \in \{3, 5, 7\} \), enabling the extraction of spatial features across varying receptive fields. This approach is critical for capturing both fine-grained details and coarse spatial patterns. Given an input feature map \({X}_f \in \mathbb{R}^{H \times W \times C} \), separate convolutional operations are performed for each kernel size \( k_i \), generating individual feature maps \({z}_i \in \mathbb{R}^{H \times W \times 1} \) as shown in \autoref{eq17}:

\begin{equation} \label{eq17}
{z}_i = \text{Conv2D}_{k_i}({X}_f) \quad \forall i \in \{1, 2, 3\}
\end{equation}

The resulting feature maps from different kernel sizes are aggregated by summation and then passed through a sigmoid activation function \( \sigma(\cdot) \), producing a spatial attention map \({A}_{\text{spatial}} \), as depicted in \autoref{eq18}:

\begin{equation} \label{eq18}{A}_{\text{spatial}} = \sigma\left( \sum_{i=1}^{3}{z}_i \right) \in \mathbb{R}^{H \times W \times 1}
\end{equation}

The spatial attention map \({A}_{\text{spatial}} \) effectively highlights spatial regions of interest by dynamically weighting the feature responses across the input dimensions. \\
\subparagraph{\textit{Frequency Domain Transformation:}}  
The 2D Fast Fourier Transform (2DFFT) is used to extract frequency-domain information, enabling the model to identify global structural patterns and regularities that might be difficult to detect in the spatial domain. For a given input feature map \({X}_f \in \mathbb{R}^{H \times W \times C} \), the 2DFFT is applied independently across the spatial dimensions \( H \) and \( W \) for each channel \( c \). This operation is represented as shown in \autoref{eq19}:
\begin{equation} \label{eq19}
{F}_{\text{freq}}(h, w, c) = \left|\text{2DFFT}({X}_f(:, :, c))\right|, \quad \forall c \in \{1, 2, \dots, C\}
\end{equation}

where \( \text{2DFFT}(\cdot) \) denotes the 2D Fast Fourier Transform, and \( |\cdot| \) represents the element-wise magnitude operation, capturing the frequency spectrum of the input features.

To integrate frequency domain information with spatial attention map, the spatial attention map \( {A}_{\text{spatial}} \in \mathbb{R}^{H \times W \times 1} \) is broadcasted along the channel dimension and element-wise multiplied with the frequency feature map \({F}_{\text{freq}} \in \mathbb{R}^{H \times W \times C} \). This operation produces a unified attention representation, as defined in \autoref{eq20}:

\begin{equation}\label{eq20}
\begin{aligned}
{A}_{\text{agg}}(h, w, c) = {A}_{\text{spatial}}(h, w,1) \odot {F}_{\text{freq}}(h, w, c), \\ \forall (h, w) \in \{1, \dots, H\} \times \{1, \dots, W\}
\end{aligned}
\end{equation}
where spatial attention map \( {A}_{\text{spatial}} \) acts as a spatial modulation mask that selectively emphasizes or suppresses specific spatial regions within the frequency domain. This modulation results in an attention-enhanced representation \( {A}_{\text{agg}} \in \mathbb{R}^{H \times W \times C} \), which effectively fuses spatial and frequency domain features. By fusing these complementary domains, the model can exploit both local spatial relationships and global frequency-based patterns, leading to a more robust and comprehensive feature representation.\\
\subparagraph{\textit{Cross-Dimensional Feature Projection:}}
The Cross-Dimensional Feature Projection of the combined attention map is achieved by applying a learnable weight matrix \( W_{\text{proj}} \) across the feature channels, followed by an element-wise multiplication with the frequency-normalized feature map \( F_{\text{norm}} \), which is derived by applying a normalization operation to the frequency-domain features \( F_{\text{freq}} \). The purpose of \( F_{\text{norm}} \) is to standardize the frequency components, ensuring they are on a consistent scale. This normalization step is critical for stabilizing the interaction between spatial and frequency-domain features, preventing the dominance of one domain over the other. This operation effectively projects the combined attention map \({A}_{\text{agg}} \in {R}^{H \times W \times C} \) into a unified spatial attention map, ${A}_{\text{proj}}$ as shown in \autoref{eq21}:
\begin{equation}\label{eq21}
{A}_{\text{proj}} = {W}_{\text{proj}} \cdot \left( {A}_{\text{agg}} \odot {F}_{\text{norm}} \right) \in \mathbb{R}^{H \times W \times C}
\end{equation}
where \({W}_{\text{proj}} \) is the learnable weight matrix, which is applied to adjust the contributions of the spatial and frequency features across all channels. This operation projects the combined features into a unified scalar form, representing the cross-channel relationship of spatial and frequency features.\\
\subparagraph{\textit{Spatial Attention Refinement:}}
To refine the spatial attention map, we introduce a spatial kernel \( {K}_{\text{spatial}} \) that captures local spatial dependencies. This step is crucial for enhancing the spatial relationships in the attention map before further processing. The spatial attention map \( {A}_{\text{proj}} \) is first modulated with the kernel \( {K}_{\text{spatial}} \) via a 2D convolution operation \( \ast \), which adjusts the attention values by considering nearby spatial information.
Subsequently, a dimensionality reduction operation is performed using a learnable weight matrix \( {W}_{\text{reduce}} \), followed by a ReLU activation function. This is followed by an expansion to the original channel dimension using another learnable weight matrix \( {W}_{\text{expand}} \). The resulting refined spatial attention map \( {A}_{\text{refined}} \) is shown in \autoref{eq23}:
\begin{equation}\label{eq23}
{A}_{\text{refined}} = \sigma\left( {W}_{\text{expand}} \left( \mathcal{R}\left( {W}_{\text{reduce}} \cdot \left( {A}_{\text{proj}} \ast {K}_{\text{spatial}} \right) \right) \right) \right)
\end{equation}
This refined attention map \( {A}_{\text{refined}} \) represents a better spatially aware attention, where the model is able to focus on regions of the feature map that are both spatially relevant and modulated by frequency-domain information. The use of the spatial kernel \( {K}_{\text{spatial}} \) ensures that local spatial dependencies are effectively captured and incorporated into the attention process, improving the model's ability to emphasize important regions of the input data.
\\
\subparagraph{\textit{Modality-Aligned Feature Recalibration:}}
The refined spatial attention map \( {A}_{\text{refined}} \) is then applied to the input feature map \( X_f \) via element-wise multiplication, along with the original spatial attention map \( {A}_{\text{Proj}} \). The final output \(Y_{\text{msa}}  \in \mathbb{R}^{H \times W \times C}\), is calculated as shown in \autoref{eq233}.
\begin{equation} \label{eq233}
Y_{\text{msa}} = \left( X_f \odot  \left( W_{\text{att}} \cdot {A}_{\text{Proj}} \odot W_{\text{refined}} \cdot {A}_{\text{refined}} \right) \right)
\end{equation}
Here, \( W_{\text{att}} \) and \( W_{\text{refined}} \) are learnable weight matrices designed to modulate the contributions of the attention mechanisms \( A_{\text{Proj}} \) and \( A_{\text{refined}} \), respectively. These weight matrices are applied to adjust the influence of each attention map, allowing the model to control the relevance of spatial features at different stages. After applying the weight matrices, element-wise multiplication (denoted by \( \odot \)) is performed to modify the contribution of the spatial attention mechanisms, enhancing the model's ability to focus on relevant features based on the learned weights. This process refines the spatial attention by allowing the model to focus on the most relevant spatial regions.\\
FMSA is a novel attention mechanism that combines spatial and frequency domain features to enhance feature representation. It uses multi-scale convolutions to capture spatial patterns at various resolutions and applies frequency modulation derived from 2DFFT to capture global structural patterns. Spatial attention emphasizes relevant regions, while frequency modulation refines global pattern detection. By fusing spatial and frequency features through element-wise operations, FMSA enables the model to capture both local spatial dependencies and global frequency patterns, improving performance in tasks requiring detailed and holistic understanding.\\
Furthermore, the feature maps produced by the FEECA and FMSA mechanisms are concatenated along the channel dimension as depicted in \autoref{ff}.
\begin{equation} \label{ff}
   Y_{\text{concat}}= \text{concat} (Y_\text{mca},Y_\text{msa}) \in \mathbb{R}^{H \times W \times C}
\end{equation}
The output feature map $Y_{\text{concat}}$ is then flattened into a vector representation, \(Y_{\text{flatten}} \in \mathbb{R}^{d'}\), where \(d' = H \times W \times C\) represents the dimensionality of the flattened feature vector. This vector is then used as input to the subsequent Attention Fusion Block, enabling the model to effectively fuse the spatial and frequency information for further processing and enhanced feature integration.
\subsubsection{Attention Fusion Block} 
The Attention Fusion Block plays a pivotal role in consolidating the outputs from different components, leading to a unified feature representation.  Finally, this flattened vector is concatenated with the global feature vector \( H_G^{\text{concat}} \)(as described in \cref{global}). This can be illustrated in \autoref{eq24}:
\begin{equation} \label{eq24}
Y_{\text{final}} = \text{concat}(Y_{\text{flatten}}, H_G^{\text{concat}}) \quad \in \mathbb{R}^{(d'+d_t+d_i)}
\end{equation}
where \( H_G^{\text{concat}} \in \mathbb{R}^{(d_t+d_i)} \) is the global feature vector.
The resulting vector, denoted as \( {Y}_{\text{final}} \), is referred to as the Global Attended Contextual Features. These features are then passed through the Global Contextual Network for further processing, providing a fusion of information from both spatially localized and global contexts. This integration not only enriches the feature representation but also optimizes the model's overall performance, enabling it to make more accurate predictions and effectively capture complex patterns within the data.
\subsection{Cascading Convolutional Transformer Feature Refinement Module}
The Cascading Convolutional Transformer Feature Refinement Module (CCTFRM) combines the strengths of gated convolutional networks, transformers, cascading connections, and adaptive scaling operations to effectively extract and refine features. This module consists of two components: (a) Cascading Convolutional Transformer and (b) Reverse Feature Harmonization. These components work together to improve semantic context and feature refinement process. 
\subsubsection{Cascading Convolutional Transformer}
The Cascading Convolutional Transformer integrates gated convolutional layers with the transformer model to efficiently capture both local and global dependencies in sequential data. Gated convolutional layers are employed to extract spatial features from the input, while transformer blocks allow for the modeling of long-range dependencies and contextual relationships. This hybrid approach enhances feature extraction and improves the model's ability to handle complex tasks, such as disaster-related event classification. The architecture comprises the following steps: (a) Input Processing and Feature Extraction, (b) Transformer Encoding, and (c) Upsampling and Cascading Connections.
\begin{figure*}
    \centering
    \includegraphics[width=\linewidth]{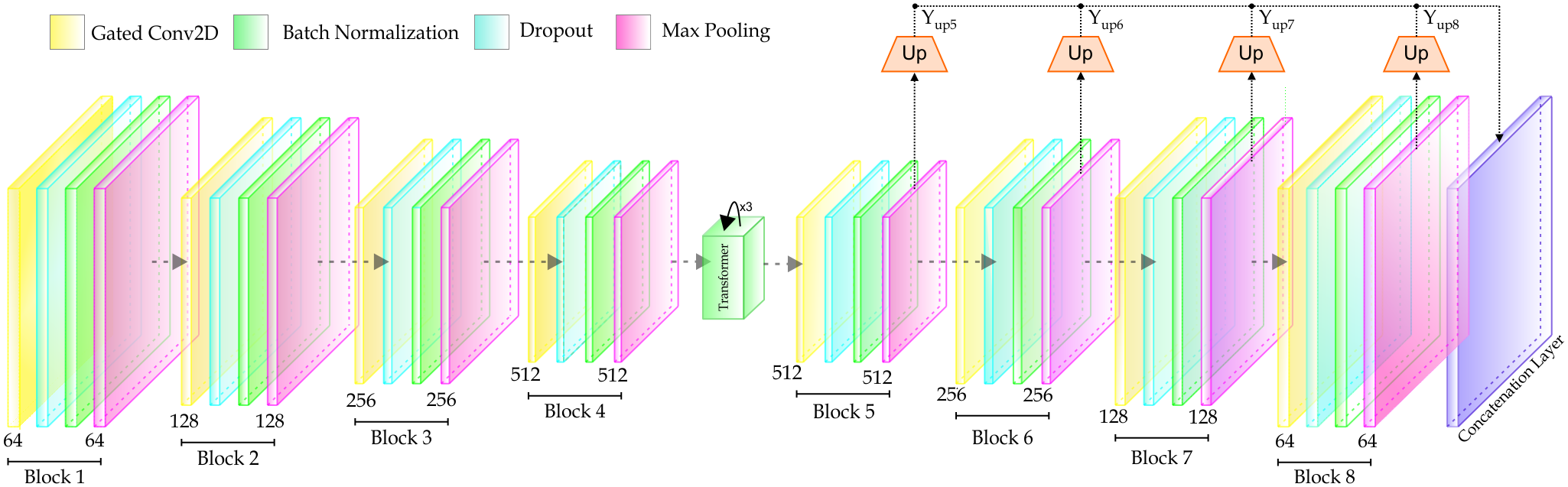}
    \caption{Diagram of the Cascading Convolutional Transformer, which integrates convolutional networks and transformers for enhanced feature extraction and refinement. The module includes key components such as the transformer encoder, gated convolutional blocks, and cascading connections. The process consists of the following steps: (a) Input Processing and Feature Extraction, (b) Transformer Encoding, and (c) Upsampling and Cascading Connections, which collectively enhance context awareness and improve the feature refinement process. \textit{Up} represents the Upsampling operation.}
    \label{CCTFRM}
\end{figure*}

\paragraph{Input Processing and Feature Extraction:}
The architecture, as illustrated in \cref{CCTFRM}, begins by processing input images through a series of downsampling blocks, each incorporating a gated convolutional layer, batch normalization, followed by dropout and max pooling. The gated convolutional block improves the feature extraction process by allowing the network to determine which features to highlight or suppress.
Let the input image \(X_i\) have dimensions \(H \times W \times C\), where \(H\) is the height, \(W\) the width, and \(C\) the number of channels. Within each block, the primary operation is a gated convolution, expressed in \autoref{eq26a} and \autoref{eq26b}:  
\begin{subequations}
\begin{align}
G &= \text{Conv2D}(X_i, K_{\text{gate}}) \label{eq26a} \\
Y_{\text{gated}}&= \ensuremath{\mathcal{M}_{pool}}\left(\mathcal{R}\left(G \odot \sigma(G)\right)\right) \label{eq26b}
\end{align}
\end{subequations}  
where \(G\) denotes the output feature map, \(K_{\text{gate}}\) the convolutional kernel of size 3 $\times$ 3, \(\sigma\) the sigmoid activation, \(\ensuremath{\mathcal{M}_{pool}}\) represents Max Pooling , and \(\odot\) represents element-wise multiplication. These steps can also be depicted in \cref{algorithm} from lines 1 to 15.
The hierarchical gating mechanism is designed to selectively amplify salient features while suppressing non-informative ones, enhancing the model's representational capacity. A sequence of downsampling blocks, such as Blocks 1 to 4, were employed, each configured with an increasing number of output channels, such as 64, 128, 256, and 512, to progressively capture multi-scale feature representations. The output feature maps from these blocks are subsequently flattened and partitioned into fixed-size patches, ensuring compatibility with the input requirements of the transformer encoder for efficient and structured feature processing.
\paragraph{Transformer Encoding:}
The reshaped patches are processed by the transformer encoder, which leverages self-attention mechanisms to capture global contextual relationships among patches, as depicted in lines 16 to 19 in \cref{algorithm}. This architecture enables the model to effectively integrate semantic information from diverse spatial and feature subspaces. With multiple transformer layers, the encoder produces a refined feature representation, which serves as the foundation for subsequent tasks. In our implementation, we employ three transformer layers to optimize feature encoding.
\paragraph{Upsampling and Cascading Connections:}
The upsampling block restores spatial dimensions while simultaneously refining features. Each upsampling block comprises an upsampling operation followed by a gated convolution, ensuring the spatial and feature-level integrity of the representations. To further enhance feature richness, cascading connections aggregate multi-scale information from different layers of the upsampling path. This cascading operation is defined as shown in \autoref{eq27}:  
\begin{equation}   \label{eq27}
Y_{\text{cascade}} = \text{Concat}(Y_{\text{up5}}, Y_{\text{up6}}, Y_{\text{up7}}, Y_{\text{up8}})
\end{equation}  
where \(Y_{\text{up5}}, Y_{\text{up6}}, Y_{\text{up7}},\) and \(Y_{\text{up8}}\) represent feature maps at successive stages of upsampling as can be seen from \cref{CCTFRM}. By combining multi-scale features, the decoder enriches the representation, enhancing the model's ability to address complex visual tasks, as shown in lines 20 to 30 in the \cref{algorithm}.

In summary, the Cascading Convolutional Transformer Feature Refinement Module integrates the hierarchical gating mechanism of the encoder with the multi-scale feature aggregation of the decoder. This synergy between gated convolutions and transformer architectures ensures optimized feature extraction and robust classification performance, representing a significant advancement in neural network design for visual tasks.

\begin{algorithm} 
\caption{Cascading Convolutional-Transformer Feature Refinement}
\label{algorithm}
\begin{algorithmic}[1]
\Require $X$: Input feature map
\Require $N$: Number of upsampling stages
\Ensure $Y_{\text{output}}$: Final output after decoding

\Function{FeatureEnhancement}{$X_{\text{in}}$}
    \State $X_{\text{up}} \gets \text{Upsample}(X_{\text{in}})$ 
    \State $X_{\text{gated}} \gets \text{GatedConv2D}(X_{\text{up}})$ 
    \State $X_{\text{dropout}} \gets \text{Dropout}(X_{\text{gated}})$ 
    \State $X_{\text{batchnorm}} \gets \text{BatchNorm}(X_{\text{dropout}})$ 
    \State $X_{\text{pooled}} \gets \text{MaxPooling}(X_{\text{batchnorm}})$ 
    \State \Return $X_{\text{pooled}}$
\EndFunction

\Procedure{Encoder}{$X$}
    \State $X_{\text{encoded}} \gets X$
    \For{$i = 1$ to $N/2$} 
        \State $X_{\text{encoded}} \gets \text{FeatureEnhancement}(X_{\text{encoded}})$
    \EndFor
    \State \Return $X_{\text{encoded}}$
\EndProcedure

\Function{Transformer}{$X_{\text{encoded}}$}
    \State $X_{\text{transformed}} \gets \text{Transformer}(X_{\text{encoded}})$ 
    \State \Return $X_{\text{transformed}}$
\EndFunction

\Procedure{Decoder}{$X, N$}
    \State $Y \gets []$ 
    \State $X_{\text{current}} \gets X$
    \For{$i = 1$ to $N/2$} 
        \State $X_{\text{current}} \gets \text{FeatureEnhancement}(X_{\text{current}})$
        \State Append $X_{\text{current}}$ to $Y$
    \EndFor
    \State \Return $Y$ 
\EndProcedure

\Function{CascadingConnections}{$Y$}
    \State $Y_{\text{cascade}} \gets \text{Concat}(Y[1], Y[2], \dots, Y[N/2])$ 
    \State \Return $Y_{\text{cascade}}$
\EndFunction

\Procedure{Main}{$X, N$}
    \State $X_{\text{encoded}} \gets \text{Encoder}(X)$ 
    \State $X_{\text{transformed}} \gets \text{Transformer}(X_{\text{encoded}})$ 
    \State $Y_{\text{upsampled}} \gets \text{Decoder}(X_{\text{transformed}}, N)$ 
    \State $Y_{\text{cascade}} \gets \text{CascadingConnections}(Y_{\text{upsampled}})$
    \State \Return $Y_{\text{cascade}}$
\EndProcedure

\end{algorithmic}
\end{algorithm}

\subsubsection{Reverse Feature Harmonization}
The Reverse Feature Harmonization layer is designed to effectively integrate and harmonize the outputs from the Cascading Convolutional Transformer Feature Refinement Module (\(Y_{\text{cascade}}\)) and image feature maps ($X_i$) by leveraging a sophisticated, adaptive fusion mechanism. This fusion mechanism enhances the interaction between the two feature sets, resulting in enhanced  representations suitable for downstream tasks such as classification.\\
The layer involves several steps, each contributing to the refinement of feature interactions. Both the feature maps are subjected to batch normalization to stabilize learning and prepare them for further fusion. The normalized feature maps are denoted as,
\(Y_\text{cascade}^{\text{N}}\) denoting the normalized Cascading Convolutional Transformer output and \(X_i^{\text{N}}\) denoting for the normalized image feature map.
The two feature maps undergo a gating subtraction operation to create a binary mask from the \(Y_\text{cascade}^{\text{N}}\) and \(X_i^{\text{N}}\) to refine their interaction. The operation is expressed in \autoref{eq28}:
\begin{equation} \label{eq28}
     Y_{\text{sub}} = \beta \cdot ( X_i^{\text{N}}) - \sigma(Y_\text{cascade}^{\text{N}})
\end{equation}
where \(\beta\) is a trainable scalar that adjusts the impact of the subtraction operation on the image features. \(Y_{\text{sub}}\) is the resulting feature after subtraction, having the same spatial dimensions as the Cascading Convolutional Transformer output, ensuring compatibility for downstream tasks. This subtraction helps modulate the image features by removing information dictated by the transformer, thereby encouraging a more dynamic interaction between the two modalities.\\
To combine the features adaptively, a gating mechanism is applied. This mechanism uses a weighted sum of the normalized output from the Cascading Convolutional Transformer and the image feature map. The gate learns how to balance the contribution of each feature set.
This is calculated by applying a weighted sum of the normalized feature maps as shown in \autoref{eq29}:
    \begin{equation} \label{eq29}
    Y_\text{gating} = \sigma\left( g_{\text{cascade}} \cdot Y_\text{cascade}^{\text{N}} + g_{\text{image}} \cdot X_i^N \right)
    \end{equation}
where \(g_{\text{cascade}}, g_{\text{image}} \) are trainable gating weights that control the importance of each feature map in the final gated output. \(Y_\text{gating}\) represents the gating output, which modulates the interaction between the Cascading Convolutional Transformer and image features.

After computing the gate, the final output \(O_{\text{final}}\) is produced by a weighted sum of the Cascading Convolutional Transformer and subtracted image features, scaled by the gating tensor \autoref{eq30}:
\begin{equation} \label{eq30}
    O_{\text{final}} = Y_\text{gating} \odot \left( \alpha_{\text{cascade}} \cdot Y_{\text{cascade}}^{\text{N}} + \alpha_{\text{sub}} \cdot Y_{\text{sub}} \right)
   \end{equation}
where \(O_{\text{final}} \in \mathbb{R}^{H_t \times W_t \times C_t}\) is the final harmonized feature map. \(\alpha_{\text{cascade}}, \alpha_{\text{sub}}\) are trainable parameters that control the contribution of each feature set to the final output.
The final output is a harmonized feature map that combines the spatial and contextual information from both the Cascading Convolutional Transformer and the image feature maps, facilitating more robust and informative representations for downstream tasks. The output is then flattened for the subsequent processes, resulting in \( O_{\text{final}} \in \mathbb{R}^{d_r} \), where \( d_r = H_t \times W_t \times C_t \).
\\
The Reverse Feature Harmonization layer serves as a crucial component by integrating feature maps from different components. Through the combination of gating mechanisms, subtraction operations, and trainable scaling parameters, the layer facilitates the adaptive fusion of features. The resulting harmonized feature map enhances the model's ability to learn complementary representations from multiple modalities, leading to better performance in tasks requiring a fine-grained understanding of the modalities.
\subsection{Unified Feature Fusion Module}
The Unified Feature Fusion Module (UFFM) is the final block in our proposed method, designed to fuse different feature maps generated by previous modules. This fusion mechanism leverages the strengths of multimodality, attention mechanisms, transformer-based context encoding, and gated convolutional feature extraction, producing a unified representation that captures comprehensive spatial and semantic relationships from all modalities. This can be expressed in \autoref{final}.
\begin{equation}\label{final}
F_{\text{concat}} = \text{concat} (Y_{\text{final}},O_{\text{final}})
\end{equation}
The input to this module includes three sets of feature maps from previous modules fused via concatenation. The unified feature vector $F_{\text{concat}} \in \mathbb{R}^{d_t+d_i+d'+d_r}$ is passed through a fully connected layer to reduce its dimensionality, followed by a sigmoid activation function to compute the final prediction probabilities. This produces the binary classification output as shown in \autoref{dense}:
\begin{equation}\label{dense}
P_{\text{f}} = \sigma(\mathcal{FC}_{\textit{L}}(F_{\text{concat}}))
\end{equation}
where $\mathcal{FC}_{\textit{L}}$ represents fully connected layers.
To optimize the model during training, we utilize the Binary Cross Entropy (BCE) loss function, which measures the error between the predicted probabilities and the ground truth labels. The BCE loss ($\mathcal{L}_{\text{BCE}}$ ) function is depicted in \autoref{eq33}:
\begin{equation} \label{eq33}
\mathcal{L}_{\text{BCE}} = -\frac{1}{N} \sum_{i=1}^{N} \left( y_i \log(P_{\text{f}, i}) + (1 - y_i) \log(1 - P_{\text{f}, i}) \right)
\end{equation}
where $y_i$ is the ground truth label, and $P_{\text{f}, i}$ represents the predicted probability for the $i$-th sample. To ensure consistent and efficient learning, the AdamW optimizer is used to update the model's parameters.\\
In summary, UFFM combines the outputs of preceding modules to produce a unified feature representation. This representation is then passed via a fully connected layer for final classification, which is optimized using binary cross-entropy loss and the AdamW optimizer. By effectively fusing output features, the UFFM significantly contributes to improving the model's overall performance.
\section{Experimental Evaluations} \label{sec4}
Our experimental evaluations validate the effectiveness of our proposed method across diverse datasets and performance metrics. We provide a thorough assessment of model performance across various evaluation metrics. In this section, we discuss the datasets employed, evaluation metrics utilized, comparison methods, experimental results, and the detailed analysis of experimental results, highlighting the strengths and advancements of our approach.
\subsection{Datasets}
The datasets used in our experiments include diverse social media data, each contributing unique characteristics to evaluate our model’s performance. Specifically, the datasets include Chennai Floods, Harz17, and Rhine18.
\subsubsection{Chennai Flood Dataset}
The Chennai Flood dataset \citep{chennai} contains a collection of social media posts, predominantly from Twitter (now known as X), gathered during the severe flooding in Chennai, India, in December 2015. It includes a total of 2,676 social media image posts, divided into 1,873 training samples, 402 testing samples, and 401 validation samples.
\subsubsection{Rhine18 Flood Dataset}
The Rhine18 Flood dataset \citep{barz} documents social media posts from the 2018 Rhine Valley flooding, providing data on flood-affected areas during the crisis. The dataset comprises a total of 1,570 samples, with 1,256 training samples, 157 testing samples, and 157 validation samples.
\subsubsection{Harz17 Flood Dataset}
The Harz17 Flood dataset \citep{barz} captures social media posts from the 2017 flood in Germany’s Harz region, providing data on flood-affected areas. It includes 487 training samples and 122 testing samples.
\subsection{Evaluation Metrics}
In our evaluation, we utilize well-established metrics, such as Accuracy (Acc), Precision (Prec), Recall (Rec), and F1-score (F1), to rigorously assess the performance of the proposed method.

\subsection{Comparision Methods}
\textbf{BERT} \citep{BERT}, developed by Google AI, is a powerful language model that is able to understand the context of the words effectively by employing a bidirectional transformer architecture.\\\\
\textbf{ViT} \citep{vit} applies the transformer model directly to image recognition by treating image patches as a sequence of tokens. It has achieved state-of-the-art performance on various image classification tasks.\\\\
\textbf{{LLaVa}} \citep{LLAVA} can understand and generate both images and text. Trained on a massive dataset of image-text pairs, LLaVA can describe images and generate images based on textual descriptions, showcasing the ability to connect visual and linguistic information.\\\\
\textbf{LlaMa} \citep{llama} is a collection of language models, ranging from 7B to 405B parameters, trained on trillions of tokens from publicly available datasets. It excels in tasks such as text generation, translation, summarization, sentiment analysis, and code generation, leveraging its transformer-based architecture. \\\\
\textbf{{Phi-3.5-Vision}} \citep{phi} released by Microsoft, trained on synthetic data, specializing in multi-frame image analysis, reasoning, multi-image summarization, and video summarization.\\\\
\textbf{InternVL} \cite{internvl} is a multimodal model designed to integrate vision and language modalities. It excels at a wide range of tasks, including image classification, video-text retrieval, image captioning, and multimodal dialogue.  \\\\
\textbf{{DMCC}} \citep{dmcc} is a multimodal method for crisis content classification that integrates textual and visual data using a two-level fusion strategy. It enhances crisis classification performance, offering valuable insights for resource allocation in disaster situations.  \\\\
\textbf{{CAMM}} \citep{camm} is a multimodal method for disaster data classification. By using cross-attention mechanisms on text and image data, CAMM captures cross-domain relationships, improving disaster classification performance.\\\\
\textbf{{RobertaMFT}} \citep{robertamft} is a multimodal fusion transformer that overcomes the limitations of traditional multimodal learning by employing a co-learning strategy. It solves data imbalance and helps to enhance recognition performance across various modalities, improving multimodal understanding and classification tasks.\\\\
\textbf{{DIF-CAM}} \citep{postfloodcbam} is a deep learning framework for classifying houses affected by floods using dual-view images. By combining aerial and ground-level images with attention mechanisms, it focuses on critical damage features, improving the accuracy of post-disaster classification.\\\\
\textbf{{UAV-CNN}} \citep{bashir2024efficient} is designed for efficient disaster event classification using UAV-captured images. By using aerial imagery and attention mechanisms, it focuses on key disaster features, thus improving classification accuracy.\\\\
\textbf{{FloodVisionNet}} \citep{yasi2024flood} is an ensemble learning approach for classifying flood and non-flood images. Integrating multiple models enhances classification accuracy, effectively identifying key features in both scenarios for reliable flood recognition and mitigation.\\\\
\textbf{{AdaptNet}} \citep{khattar2022generalization} generalizes convolutional neural networks for domain adaptation, enabling the classification of disaster-related images. By adapting to varying image distributions, model's ability to identify disaster events across diverse contexts gets enhanced.\\\\
\subsection{Experimental Results}
In this section, we provide a detailed analysis of the experimental setup, network hyperparameters, performance comparisons, and ablation studies.
\subsubsection{Experimental Setup}
All experiments are conducted on a Linux server equipped with an Intel(R) Xeon(R) Silver 4215R CPU @3.20 GHz, 256GB of RAM, and an NVIDIA Tesla T4 GPU. The XFloodNet framework was implemented using TensorFlow, and some state-of-the-art methods are implemented in Pytorch, leveraging the Hugging Face Transformers library for pre-trained language and vision models. The AdamW optimizer was employed for model training with a learning rate of 10$^{-4}$. A dropout rate of 0.2 was applied to prevent overfitting. Models were trained for 100 epochs. This setup provides a robust and reproducible experimental environment for evaluating the performance of XFloodNet.
 \subsubsection{Network Hyperparameters}
This section presents a hyperparameter optimization for our proposed method, addressing critical parameters including learning rate, number of epochs,  batch size, optimizer and dropout rate selection to enhance overall performance. The learning rate was systematically explored in the range \(10^{-3}\) to \(10^{-6}\), with \(10^{-4}\) identified as the optimal value for convergence. Epoch values were tested at 30, 50, 100, and 120, with 100 epochs providing the best trade-off between convergence speed and performance. For batch size, we evaluated 8, 16,32, and 64 and found 32 to be the most efficient in terms of computational resources. Dropout rates between 0.1 and 0.4 were assessed, where a rate of 0.2 yielded the most effective regularization. After comparing multiple optimization algorithms, AdamW was selected for its superior ability to handle weight decay and improve model generalization. The final optimized configuration as shown in \cref{Table Hyperparameters} a learning rate of \(10^{-4}\), 32 as batch size, 100 as the number of epochs, dropout rate as 0.2, and AdamW optimizer represents the outcome of exhaustive experimentation, ensuring the model performs at its best.
\begin{table}[pos=ht] 
\caption{Hyperparameters Used in XFloodNet} \label{Table Hyperparameters}
\centering
\begin{tabular}{l@{\hspace{3.00cm}}l}
\toprule
\textbf{Hyperparameter} & \textbf{Value} \\
\midrule
Learning Rate           & 0.0001          \\
Batch Size              & 32             \\
Optimizer               & AdamW          \\
Dropout Rate            & 0.2           \\
Number of Epochs        & 100            \\
Activation Function     & ReLU           \\
Loss Function           & Cross-Entropy  \\
Weight Decay  & 0.0001       \\
\bottomrule
\end{tabular}
\end{table}
\subsubsection{Comparative Performance Assessment on the Chennai Flood Dataset}
The performance comparison \cref{Chennai} provides a detailed view of how different models, spanning pretrained baselines, vision-language models, state-of-the-art methods, and our proposed method XFloodNet, perform on the Chennai Flood dataset. The evaluation metrics such as accuracy (Acc), precision (Prec), recall (Rec), and F1-score (F1) offer insight into each model’s ability for classification performance across flood-related data. The symbol $\boldsymbol{\Delta}$ represents the absolute difference in model performance between the proposed method and the corresponding method across all evaluation parameters.\\
\begin{table*}[ht]
    \centering
    \caption{Comparative Performance Assessment on Chennai Flood Dataset}
    \begin{tabular}{l@{\hspace{0.5cm}}l@{\hspace{0.5cm}}l@{\hspace{0.5cm}}l@{\hspace{0.5cm}}l@{\hspace{0.5cm}}l@{\hspace{0.5cm}}l@{\hspace{0.5cm}}l@{\hspace{0.5cm}}l@{\hspace{0.5cm}}l}
        \toprule
        \textbf{Category} & \textbf{Methods} & \textbf{Acc} & \boldmath$\Delta_{\text{Acc}}$ & \textbf{Prec} & \boldmath$\Delta_{\text{Prec}}$ & \textbf{Rec} & \boldmath$\Delta_{\text{Rec}}$ & \textbf{F1} & \boldmath$\Delta_{\text{F1}}$ \\
        \midrule
        \multirow{2}{*}{\textbf{Pretrained Baseline Models}} 
          & BERT \citep{BERT}   & 88.27 & 7.74  & 89.16 & 4.95  & 75.88 & 16.68  & 81.99 & 11.34  \\
          & ViT \citep{vit}  & 86.53 & 9.48  & 87.17 & 6.94  & 72.34 & 20.22  & 79.06 & 14.27  \\
        \midrule
        \multirow{4}{*}{\textbf{Vision-Language Models}}
          & LLaMA \citep{llama}& 45.56 & 50.45 & 35.32 & 58.79 & 76.72 & 15.84  & 48.37 & 44.96  \\
          & LLaVA \citep{LLAVA} & 69.02 & 26.99 & 50.87 & 43.24 & \textbf{98.48} & 5.92  & 67.09 & 26.24  \\
          & Phi-3.5-Vision \citep{phi}& 49.86 & 46.15 & 33.52 & 60.59 & 51.72 & 40.84  & 40.68 & 52.65  \\
          & InternVL \citep{internvl} & 72.89 & 23.12 & 53.03 & 41.08 & 86.78 & 5.78   & 65.83 & 27.50  \\
        \midrule
        \multirow{8}{*}{\textbf{State-of-the-art Methods}}
          & DIF-CAM  \citep{postfloodcbam}   & 73.38 & 22.63 & 81.82 & 12.29 & 73.38 & 19.18  & 77.37 & 15.96  \\
          & UAV-CNN \citep{bashir2024efficient}        & 69.90 & 26.11 & 48.86 & 45.25 & 69.90 & 22.66  & 57.52 & 35.81  \\
          & FloodVisionNet \citep{yasi2024flood} & 90.80 & 5.21 & 90.72 & 3.39 & 73.62 & 18.94  & 81.28 & 12.05  \\
          & AdaptNet \citep{khattar2022generalization} & 80.11 & 15.90 & 81.10 & 13.01 & 83.00 & 9.56   & 82.74 & 10.59   \\
          & DMCC \citep{dmcc} & 91.26 & 4.75  & 92.53 & 1.58  & 82.81 & 9.75   & 87.40 & 5.93   \\
          & CAMM \citep{camm} & 91.51 & 4.50  & 92.49 & 1.62  & 82.51 & 10.05  & 87.21 & 6.12   \\
          & RobertaMFT \citep{robertamft}& 89.27 & 6.74  & 86.02 & 8.09  & 82.97 & 9.59   & 84.47 & 8.86   \\
        \midrule
        \textbf{Proposed Method} &\textbf{ XFloodNet} & \textbf{96.01} & 0.00    & \textbf{94.11} & 0.00    & {92.56} & 0.00     & \textbf{93.33} & 0.00     \\
        \bottomrule
    \end{tabular}
    \label{Chennai}
\end{table*}
We evaluate the performance of pre-trained baseline models, BERT and ViT, employed in our proposed approach. BERT achieves an Acc of 88.27\%, with a Prec of 89.16\%, a Rec of 75.88\%, and an F1 of 81.99\%. These metrics indicate a balanced performance in identifying relevant flood-related textual content. ViT exhibits improved performance with an Acc of 86.53\%, Prec of 87.17\%, Rec of 72.34\%, and F1 of 79.06\%. The strong performance of BERT and ViT can be attributed to their ability to effectively capture textual and visual features, respectively, leveraging pre-trained knowledge to accurately identify flood-related content and imagery. \\
Next, we compare several Vision Language Models, such as LLaMA \citep{llama}, LLaVA \citep{LLAVA}, InternVL \citep{internvl}, and Phi-3.5-Vision \citep{phi}. LLaMA achieves an Acc of 45.56\%, with a Prec of 35.32\% and a Rec of 76.72\%, resulting in an F1 of 48.37\%. This indicates a reasonable ability to identify relevant content, but it falls short compared to the pre-trained baseline models. Despite LLaMA being trained on a vast corpus of data, its general-purpose nature and lack of domain-specific fine-tuning for flood-related tasks limit its ability to effectively capture and classify relevant flood-related content. LLaVA presents an Acc of 69.02\%, paired with a Prec score of 50.87\%, with high Rec score of 98.48\%, and an F1 of 67.09\%. LLaVA’s higher recall indicates its greater sensitivity in detecting relevant instances, even if it leads to more false positives, highlighting its ability to capture a broader range of flood-related content. XFloodNet outperforms LLaVA in Acc, Prec, and F1 due to its optimized architecture for better discrimination between relevant and non-relevant flood content. Phi-3.5-Vision shows an Acc of 49.86\%, a Prec of 33.52\%, a Rec of 51.72\%, and an F1 of 40.68\%. Phi-3-Vision's lower performance can be attributed to its relatively limited ability to capture complex flood-related patterns, as evidenced by its lower Prec and F1 scores. This indicates that while Phi can identify some relevant content, it struggles with distinguishing non-relevant content, leading to a higher number of false positives. Additionally, its relatively lower Prec suggests that the model might be more prone to misclassify non-flood-related instances as relevant. InternVL reports $72.89\%$ $\text{Acc}$, $53.03\%$ $\text{Prec}$, $86.78\%$ $\text{Rec}$, and $65.83\%$ $\text{F1}$, showcasing high recall at the expense of lower precision. \\
The analysis encompasses seven state-of-the-art models, namely DIF-CAM \citep{postfloodcbam}, UAV-CNN \citep{bashir2024efficient}, FloodVisionNet \citep{yasi2024flood}, AdaptNet \citep{khattar2022generalization}, DMCC \citep{dmcc}, CAMM \citep{camm}, and RobertaMFT \citep{robertamft}. 
DIF-CAM achieves a balanced performance with $73.38\%$ $\text{Acc}$, $81.82\%$ $\text{Prec}$, $73.38\%$ $\text{Rec}$, and $77.37\%$ $\text{F1}$. The relatively lower performance of DIF-CAM compared to XFloodNet can be attributed to its reliance on the dual-view convolutional neural network framework, which, while effective at integrating interior and exterior visual indicators, lacks advanced multimodal fusion capabilities. XFloodNet incorporates state-of-the-art architectures like ViT and BERT, which effectively combine spatial, textual, and contextual features, thereby leveraging a more comprehensive representation of the data.  
UAV-CNN, in contrast, demonstrates $69.90\%$ $\text{Acc}$, $48.86\%$ $\text{Prec}$, $69.90\%$ $\text{Rec}$, and $57.52\%$ $\text{F1}$, reflecting a significant drop in precision despite maintaining comparable accuracy and recall. UAV-CNN's low performance than XFloodNet can be attributed to its reliance on basic CNN architectures, which struggle to capture complex spatial and contextual relationships in flood images. Additionally, the lack of advanced attention mechanisms and multimodal fusion limits its ability to distinguish fine differences between flood-related images.
FloodVisionNet achieves $90.80\%$ $\text{Acc}$, $90.72\%$ $\text{Prec}$, $73.62\%$ $\text{Rec}$, and $81.28\%$ $\text{F1}$, balancing precision and recall. FloodVisionNet may exhibit lower performance in certain scenarios due to potential issues with model diversity and overfitting in the ensemble. While combining multiple CNN architectures enhances generalization, it can also introduce redundancy, where models with similar feature extraction capabilities may not contribute significantly to improving accuracy. Additionally, if the ensemble lacks proper weight balancing or model selection during training, it could lead to suboptimal performance, especially if certain models dominate the final decision-making process. The complexity of managing diverse models can also increase computational requirements, potentially limiting its ability to handle more intricate, unseen flood scenarios effectively. \\
AdaptNet demonstrates strong performance with $80.11\%$ $\text{Acc}$, $81.10\%$ $\text{Prec}$, $83.00\%$ $\text{Rec}$, and $82.74\%$ $\text{F1}$. AdaptNet demonstrates lower performance than XFloodNet due to its dependence on Maximum Mean Discrepancy (MMD) for aligning source and target domain distributions, which may struggle to capture complex relationships that are inherent in flood-related data.
The performance metrics of DMCC, CAMM, and RobertaMFT further represent the variance that is present among the models. DMCC outperformed with $91.26\%$ $\text{Acc}$, $92.53\%$ $\text{Prec}$, $82.81\%$ $\text{Rec}$, and $87.40\%$ $\text{F1}$. DMCC employs an adaptive attention mechanism within the Multimodal Channel Attention block to dynamically weigh the importance of each modality, enhancing flood categorization.\\
CAMM recorded $91.51\%$ $\text{Acc}$, $92.49\%$ $\text{Prec}$, $82.51\%$ $\text{Rec}$, and $87.21\%$ $\text{F1}$. CAMM employs Bi-LSTMs for temporal dependencies in textual data and VGG-16 for extracting spatial features from images. However, its requirement for complete input sequences limits its applicability in low-latency, real-time environments.
RobertaMFT achieved $89.27\%$ $\text{Acc}$, $86.02\%$ $\text{Prec}$, $82.97\%$ $\text{Rec}$, and $84.47\%$ $\text{F1}$. 
In comparison, XFloodNet excels by leveraging advanced multimodal fusion strategies and feature representation through architectures like ViT and BERT. This enables it to process both spatial and textual cues effectively, surpassing these state-of-the-art models in classification performance metrics. The inferior performance of competing models is largely attributed to their restricted data representations, highlighting the necessity for robust multimodal approaches as implemented in XFloodNet.
\begin{table*}[pos=ht]
  \caption{Comparative Performance Assessment on Rhine18 Flood Dataset} \label{Table rhine}

  \begin{tabular}{l@{\hspace{0.4cm}}l@{\hspace{0.4cm}}l@{\hspace{0.4cm}}l@{\hspace{0.4cm}}l@{\hspace{0.4cm}}l@{\hspace{0.4cm}}l@{\hspace{0.4cm}}l@{\hspace{0.4cm}}l@{\hspace{0.4cm}}l}
    \toprule
      & \textbf{Methods} & \textbf{Acc} & \boldmath$\Delta_{\text{Acc}}$ & \textbf{Prec} & \boldmath$\Delta_{\text{Prec}}$ & \textbf{Rec} & \boldmath$\Delta_{\text{Rec}}$ & \textbf{F1} & \boldmath$\Delta_{\text{F1}}$ \\
    \midrule
    \multirow{2}{*}{\textbf{Pretrained Baseline Models}}
      & BERT\citep {BERT}  & 74.46 & 10.86 & 75.32 & 6.78 & 76.24 & 6.16 & 72.72 & 9.52 \\
      & ViT\citep {vit} & 71.06 & 14.26 & 59.28 & 22.82 & 88.29 & 5.89 & 70.94 & 11.30 \\
    \midrule
    \multirow{4}{*}{\textbf{Vision-Language Models}}
      & LLaMA \citep{llama}  & 44.92 & 40.40 & 37.36 & 44.74 & 75.58 & 6.82 & 50.00 & 32.24 \\
      & LLAVA \citep{LLAVA} & 63.29 & 22.03 & 52.76 & 29.34 & 92.95 & 10.55 & 67.31 & 14.93 \\
      & Phi-3.5-Vision \citep{phi} & 41.53 & 43.79 & 34.34 & 47.76 & 66.28 & 16.12 & 45.24 & 37.00 \\
      & InternVL \citep{internvl} & 65.25 & 20.07 & 51.25 & 30.85 & \textbf{95.35} & 12.95 & 66.67 & 15.57 \\
    \midrule
    \multirow{8}{*}{\textbf{State-of-the-art Methods}} 
      & DIF-CAM \citep{postfloodcbam}  & 64.83 & 20.49 & 63.64 & 18.46 & 64.83 & 17.57 & 54.75 & 27.49 \\
      & UAV-CNN \citep{bashir2024efficient}  & 63.56 & 21.76 & 40.40 & 41.70 & 63.56 & 18.84 & 49.40 & 32.84 \\
      & FloodVisionNet \citep{yasi2024flood}  & 66.80 & 18.52 & 64.90 & 17.20 & 62.60 & 19.80 & 63.71 & 18.53 \\
      & AdaptNet \citep{khattar2022generalization}  & 65.16 & 20.16 & 60.43 & 21.67 & 63.56 & 18.84 & 59.40 & 22.84 \\
      & DMCC \citep{dmcc}  & 80.85 & 4.47 & 73.33 & 8.77 & 81.91 & 0.49 & 77.38 & 14.37 \\
      & CAMM \citep{camm}  & 78.72 & 6.60 & 79.15 & 2.95 & 78.72 & 3.68 & 78.93 & 3.31 \\
      & RobertaMFT \citep{robertamft}  & 80.42 & 4.90 & 74.48 & 7.62 & 77.65 & 4.75 & 76.04 & 6.20 \\
    \midrule
       & \textbf{XFloodNet} & \textbf{85.32} & 0.00 & \textbf{82.10} & 0.00 & {82.40} & 0.00 & \textbf{82.24} & 0.00 \\
    \bottomrule
  \end{tabular}
\end{table*}
\subsubsection{Comparative Performance Assessment on Rhine18 Flood Dataset}
The performance analysis presented in \cref{Table rhine} offers a comprehensive comparison of various methodologies applied to the Rhine18 dataset. This includes pre-trained baseline models, vision-language architectures, state-of-the-art techniques, and the proposed XFloodNet framework. The comparison rigorously evaluates each approach in terms of key performance metrics, providing a systematic assessment of their capabilities. XFloodNet demonstrates superior performance, showcasing its robustness and effectiveness in addressing the specific challenges inherent to the Rhine dataset.
XFloodNet achieves an Acc of 85.32\%, Prec of 82.10\%, Rec of 82.40\%, and F1 of 82.24\%. XFloodNet outperforms models such as BERT, ViT, LLaMA, LLAVA, and Phi-3.5-Vision due to its hierarchical cross-modal gated attention, which enables deep interaction between textual and visual features, and its Heterogeneous Convolution Adaptive Multi-scale Attention Module (HCAMAM), which combines group and point convolutions for refined spatial and pixel-level feature extraction and interaction for capturing local and global features important for the flood detection. InternVL achieves high recall due to its extensive contrastive training on noisy, web-scale image-text pairs, which enables broad generalization across diverse data. The generative training with high-quality filtered data further refines cross-modal feature alignment, improving recall in complex scenarios. \\
Additionally, the supervised fine-tuning with high-quality instruction data enhances the model's ability to capture subtle, relevant patterns across modalities, reducing false negatives.
The state-of-the-art methods such as DIF-CAM, UAV-CNN, FloodVisionNet, AdaptNet, DMCC, CAMM, and RobertaMFT show lower performance than XFloodNet due to several key factors. XFloodNet incorporates several novel techniques, such as hierarchical feature interaction and cascading convolutional refinement, which are tailored to handle the unique challenges of flood-related data. Furthermore, its Heterogeneous Convolution Adaptive Multi-scale Attention Module ensures global and local feature extraction and interaction, especially for complex, multi-scale flood patterns, compared to methods such as  UAV-CNN or CAMM, which lack such specialized attention strategies. These advanced architectural features enable XFloodNet to address the specific challenges of flood detection better, leading to its superior performance over these other state-of-the-art methods.
\begin{table*}[pos=ht]
\centering
\caption{Comparative Performance Assessment on Harz17 Flood Dataset}
\label{Table harz}
\begin{tabular}{l@{\hspace{0.4cm}}l@{\hspace{0.4cm}}c@{\hspace{0.4cm}}c@{\hspace{0.4cm}}c@{\hspace{0.4cm}}c@{\hspace{0.4cm}}c@{\hspace{0.4cm}}c@{\hspace{0.4cm}}c@{\hspace{0.4cm}}c}
\toprule
 & \textbf{Methods} & \textbf{Acc} & \boldmath$\Delta_{\text{Acc}}$ & \textbf{Prec} & \boldmath$\Delta_{\text{Prec}}$ & \textbf{Rec} & \boldmath$\Delta_{\text{Rec}}$ & \textbf{F1} & \boldmath$\Delta_{\text{F1}}$ \\
\midrule
\multirow{2}{*}{Pretrained Baseline Models}
 & BERT \citep{BERT} & 79.12 & 12.38 & 81.81 & 7.19 & 54.54 & 33.66 & 65.45 & 23.15 \\
 & ViT \citep{vit} & 72.52 & 18.98 & 60.00 & 29.00 & 72.72 & 15.48 & 65.75 & 22.85 \\
\midrule
\multirow{4}{*}{Vision-Language Models}
 & LLaMA \citep{llama} & 43.48 & 48.02 & 40.00 & 49.00 & 66.67 & 21.53 & 50.00 & 38.60 \\
 & LLAVA \citep{LLAVA} & 68.64 & 22.86 & 55.61 & 33.39 & \textbf{97.90} & 9.70 & 70.93 & 17.67 \\
 & Phi-3.5-Vision \citep{phi} & 42.39 & 49.11 & 37.93 & 51.07 & 56.41 & 31.79 & 45.36 & 43.24 \\
 & InternVL \citep{internvl} & 65.25 & 26.25 & 51.25 & 37.75 & 95.35 & 7.15 & 66.67 & 21.93 \\
\midrule
\multirow{8}{*}{State-of-the-art Methods}
 & DIF-CAM \citep{postfloodcbam} & 59.78 & 31.72 & 75.00 & 14.00 & 59.78 & 28.42 & 48.41 & 40.19 \\
 & UAV-CNN \citep{bashir2024efficient} & 57.61 & 33.89 & 33.19 & 55.81 & 57.61 & 30.59 & 42.11 & 46.49 \\
 & FloodVisionNet \citep{yasi2024flood} & 64.60 & 26.90 & 70.90 & 18.10 & 61.50 & 26.70 & 51.20 & 37.40 \\
 & AdaptNet \citep{khattar2022generalization} & 54.35 & 37.15 & 47.37 & 41.63 & 69.23 & 18.97 & 56.25 & 32.35 \\
 & DMCC \citep{dmcc} & 85.71 & 5.79 & 85.63 & 3.37 & 85.71 & 2.49 & 85.66 & 2.94 \\
 & CAMM \citep{camm} & 85.10 & 6.40 & 81.20 & 7.80 & 82.30 & 5.90 & 81.75 & 6.85 \\
 & RobertaMFT \citep{robertamft} & 84.61 & 6.89 & 80.64 & 8.36 & 75.75 & 12.45 & 78.12 & 10.48 \\
\midrule
 & \textbf{XFloodNet} & \textbf{91.50} & \textbf{0.00} & \textbf{89.00} & \textbf{0.00} & {88.20} & \textbf{0.00} & \textbf{88.60} & \textbf{0.00} \\
\bottomrule
\end{tabular}
\end{table*}

\subsubsection{Comparative Performance Assessment on Harz17 Flood Dataset}
The performance analysis in \cref{Table harz} offers an evaluation of multiple methods, encompassing pre-trained baselines, vision-language models, state-of-the-art methods, and our proposed XFloodNet, all tested on the Harz17 Flood dataset. XFloodNet achieves an Acc of 91.50\%, Prec of 89.00\%, Rec of 88.20\%, and F1 of 88.60\%. XFloodNet outperforms all the methods, such as BERT, ViT, LLaMA, Phi-3.5-Vision, DMCC, and CAMM, across all the performance metrics. LLaVA’s higher recall reflects its ability to identify a wider range of relevant instances, even at the expense of increased false positives, which demonstrates its capacity to capture a broader spectrum of flood-related information. However, XFloodNet surpasses LLaVA in Acc, Prec, and F1. Moreover, XFloodNet outperforms other models due to various techniques employed in the architecture that synergize across modalities. The Multimodal Feature Interaction Module employed in the architecture of the proposed approach effectively leverages zero-shot prompting and hierarchical gated attention to capture meaningful text-visual correlations. The Heterogeneous Convolution Adaptive Multi-scale Attention Module combines group and point convolutions with multi-scale attention, enabling fine-grained and global feature integration while preserving critical information via residual connections. Additionally, the Cascading Convolutional Transformer Feature Refinement Module enhances class-specific features and ensures contextual harmonization through the Reverse Feature Harmonizer. These specific optimizations surpass the generalized capabilities of baseline, vision-language models, and state-of-the-art methods.
\subsubsection{Model Sensitivity Analysis}
We evaluate XFloodNet's performance enhancements using different combinations of modalities, modules, and attention mechanisms on the Chennai Flood dataset. First, we explore XFloodNet's performance enhancements using a variety of modality combinations. Second, we investigate the performance gain achieved by various components in the XFloodNet. Finally, we explore how different attention combinations employed in the proposed approach affect the model's overall effectiveness. 

\paragraph{Analysis on the basis of modality:}
In this section, we assess the effect of each modality employed in the XFloodNet. Considering the unique attributes of each modality, it is essential to analyze how different combinations influence the performance of our approach. \cref{modality_wise} displays the performance outcomes for these various modality combinations.\\
\begin{table*}[pos=ht]
  \caption{Performance Gain Analysis on Modalities} \label{modality_wise}

  \begin{tabular}{l@{\hspace{0.8cm}}l@{\hspace{0.8cm}}l@{\hspace{0.8cm}}l@{\hspace{0.8cm}}l@{\hspace{0.8cm}}l@{\hspace{0.8cm}}l@{\hspace{0.8cm}}l@{\hspace{0.8cm}}l}
    \toprule
    \textbf{Methods} & \textbf{Acc} & \boldmath$\Delta_{\text{Acc}}$ & \textbf{Prec} & \boldmath$\Delta_{\text{Prec}}$ & \textbf{Rec} & \boldmath$\Delta_{\text{Rec}}$ & \textbf{F1} & \boldmath$\Delta_{\text{F1}}$ \\
    \midrule
    XFloodNet (w/o Image) & 87.56 & 8.45 & 73.50 & 20.61 & 91.73 & 0.83 & 81.61 & 11.72 \\
    XFloodNet (w/o Text)  & 91.04 & 4.97 & 81.95 & 12.16 & 90.08 & 2.48 & 85.82 & 7.51 \\
    \midrule
      \textbf{XFloodNet} & \textbf{96.01} & 0.00 & \textbf{94.11} & 0.00 & \textbf{92.56} & 0.00 & \textbf{93.33} & 0.00 \\
    \bottomrule
  \end{tabular}
\end{table*}
XFloodNet (w/o Image) evaluates the importance of visual input in XFloodNet by removing the image data while retaining text features. The absence of image data hinders the model of important visual context, which reduces its ability to fully understand flood-related information. Also, the model does not utilize the complementary insights gained from text-image integration.
Furthermore, the application of attention mechanisms intended to focus on pertinent information across modalities is limited by the absence of visual modality. The advantages of CCTFRM and HCAMAM are not conceded to this framework. As a result, XFloodNet (w/o Image) has trouble matching the performance of the original model, which accounts for all data modalities. \\
XFloodNet (w/o Text), on the other hand, examines the function of textual data by removing text while retaining image features. With this configuration, we observe how textual data helps the model to understand and categorize flood-related content. There are several problems associated with this version. The absence of textual data limits the model's semantic richness and cross-modal interactions. Consequently, XFloodNet (w/o Text) may face challenges in comprehensively capturing flood-related content.
\paragraph{Analysis on the basis of module-wise components:} 
This ablation study provides a comprehensive evaluation of each module's contribution to the XFloodNet method. By systematically removing one or more modules and measuring the resulting impact on performance metrics. \cref{module_wise} highlighting the changes in performance when modules are individually or jointly omitted.
XFloodNet achieves high Acc (96.01\%), Prec (94.11\%), Rec (92.56\%), and F1 (93.33\%), showcasing the collective effectiveness of all integrated modules. This indicates that the proposed method can reliably detect flood-related content, capturing both positive and negative instances effectively due to the synergy among the modules. \\
\begin{table*}[ht]
  \caption{Performance Gain Analysis on Model Components} \label{module_wise}
  \centering
  \begin{tabular}{l@{\hspace{0.5cm}}l@{\hspace{0.5cm}}l@{\hspace{0.5cm}}l@{\hspace{0.5cm}}l@{\hspace{0.5cm}}l@{\hspace{0.5cm}}l@{\hspace{0.5cm}}l@{\hspace{0.5cm}}l}
    \toprule
    \textbf{Methods} & \textbf{Acc} & \boldmath$\Delta_{\text{Acc}}$ & \textbf{Prec} & \boldmath$\Delta_{\text{Prec}}$ & \textbf{Rec} & \boldmath$\Delta_{\text{Rec}}$ & \textbf{F1} & \boldmath$\Delta_{\text{F1}}$ \\
    \midrule
    XFloodNet (w/o MFIM)      & 93.32 & 2.69 & 92.25 & 1.86 & 91.13 & 1.43 & 91.69 & 1.64 \\
    XFloodNet (w/o HCAMAM)     & 93.56 & 2.45 & 92.32 & 1.79 & 91.78 & 0.78 & 91.05 & 2.28 \\
    XFloodNet (w/o CCTFRM)     & 94.05 & 1.96 & 91.03 & 3.08 & 90.50 & 2.06 & 90.76 & 2.57 \\
    XFloodNet (w/o (MFIM and HCAMAM))  & 90.78 & 5.23 & 88.45 & 5.66 & 89.24 & 3.32 & 88.05 & 5.28 \\
    XFloodNet (w/o (MFIM and CCTFRM))  & 91.54 & 4.47 & 89.10 & 5.01 & 89.78 & 2.78 & 89.43 & 3.90 \\
    XFloodNet (w/o (HCAMAM and CCTFRM)) & 92.89 & 3.12 & 90.80 & 3.31 & 91.11 & 1.45 & 90.95 & 2.38 \\
    \midrule
    \textbf{XFloodNet} & \textbf{96.01} & \textbf{0.00} & \textbf{94.11} & \textbf{0.00} & \textbf{92.56} & \textbf{0.00} & \textbf{93.33} & \textbf{0.00} \\
    \bottomrule
  \end{tabular}
\end{table*}
XFloodNet (w/o MFIM): When MFIM is removed, Acc drops to 93.32\%, Prec to 92.25\%, Rec to 91.13\%, and F1 to 91.69\%. The reduction in accuracy and F1-score implies that MFIM is crucial in enhancing the model's overall classification performance, suggesting it may play a role in improving the model’s decision boundary or in processing certain features critical for correct classification.\\
{XFloodNet (w/o HCAMAM):} The performance of the XFloodNet variant without HCAMAM highlights its critical importance, as all evaluation metrics decline, such as {Acc} drops to 93.56\%, {Prec} to 92.32\%, while {Rec} remains relatively high at 91.78\%, and {F1} decreases to 91.05\%. These results suggest that the model's ability to utilize frequency-domain information and capture both local and global cross-dimensional interactions is reduced when HCAMAM is excluded.\\
XFloodNet (w/o CCTFRM): XFloodNet's performance declines when the CCTFRM module is removed, Acc drops to 94.50\%, Prec declines to 91.03\%, Rec to 90.50\%, and F1 reduces to 90.76\%. The model gets deprived of the benefits of gated convolutional interactions and Reverse Feature Harmonization, which is essential for accurately capturing true positive instances.  The semantic features of data and the challenges with the diverse nature of the flood-related content can not be addressed due to the absence of these modules. \\
{XFloodNet (w/o (MFIM and HCAMAM)):} This variant of the proposed method is without the key modules such as MFIM and HCAMAM. Without them a significant decline in performance metrics is observed.  Acc decreases to 90.78\%, Prec falls to 88.45\%, Rec drops to 89.24\%, and F1 reduces to 88.05\%. This combined negative impact shows that MFIM and HCAMAM work together to help the model correctly identify true positives while reducing errors and misclassifications. In this variant, the model is deprived of intra-modality, cross-modal, local and global context, frequency domain features, and cross-dimensional interactions, which is important for the classification performance of the model.\\
{XFloodNet (w/o (MFIM and CCTFRM)):} The absence of both MFIM and CCTFRM reduces Acc to 91.54\%, Prec to 89.10\%, Rec to 89.78\%, and F1 to 89.43\%. The removal of these modules disrupts the balance, limiting its ability to detect relevant instances accurately. This analysis highlights MFIM’s and CCTFRM’s critical role in boosting classification performance and demonstrates their combined contribution to the model’s overall effectiveness in identifying flood-related content. \\
{XFloodNet (w/o (HCAMAM and CCTFRM)):} This version of XFloodNet without the HCAMAM and CCTFRM module shows an Acc drop to 92.89\%, Prec to 90.80\%, Rec to 91.11\%, and F1 to 90.95\% as compared with the proposed framework. This highlights the importance of these components towards the performance metrics in the proposed method. They are crucial for maintaining XFloodNet's balanced performance across all metrics, ensuring effective feature representation and robust detection of relevant instances. \\
This ablation study shows the importance of each module within our framework. The various variants of XFloodNet show less performance than the XFloodNet. This exhibits the importance of all modules and why their integration yields the best performance across all parameters.
\paragraph{Analysis on the basis of attention mechanisms:}
\begin{table*}[ht]
  \centering
  \caption{Performance Gain Analysis on Attention Mechanisms} \label{attention_wise}
  \begin{tabular}{l@{\hspace{0.5cm}}l@{\hspace{0.5cm}}l@{\hspace{0.5cm}}l@{\hspace{0.5cm}}l@{\hspace{0.5cm}}l@{\hspace{0.5cm}}l@{\hspace{0.5cm}}l@{\hspace{0.5cm}}l}
    \toprule
    \textbf{Methods} & \textbf{Acc} & \boldmath$\Delta_{\text{Acc}}$ & \textbf{Prec} & \boldmath$\Delta_{\text{Prec}}$ & \textbf{Rec} & \boldmath$\Delta_{\text{Rec}}$ & \textbf{F1} & \boldmath$\Delta_{\text{F1}}$ \\
    \midrule
    XFloodNet (w/o HCGAM)       & 94.13 & 1.88 & 91.50 & 2.61 & 89.56 & 2.00 & 90.51 & 2.82 \\
    XFloodNet (w/o FEECA)       & 95.12 & 0.89 & 92.80 & 1.31 & 91.00 & 1.56 & 91.64 & 1.69 \\
    
    XFloodNet (w/o FMSA)     & 94.92 & 1.09 & 92.50 & 2.61 & 91.40 & 1.16 & 91.45 & 1.88 \\
    
    XFloodNet (w/o (HCGAM and FEECA)) & 93.83 & 2.18 & 89.10 & 4.01 & 90.10 & 3.23 & 89.59 & 3.74 \\
    XFloodNet (w/o (HCGAM and FMSA)) & 92.72 & 3.29 & 90.40 & 3.71 & 90.90 & 1.66 & 90.65 & 2.68 \\
    XFloodNet (w/o (FMSA and FEECA)) & 92.90 & 3.11 & 91.00 & 3.11 & 90.10 & 2.46 & 90.55 & 2.78 \\
    \midrule
    \textbf{XFloodNet} & \textbf{96.01} & \textbf{0.00} & \textbf{94.11} & \textbf{0.00} & \textbf{92.56} & \textbf{0.00} & \textbf{93.33} & \textbf{0.00} \\
    \bottomrule
  \end{tabular}
\end{table*}
This ablation study analyzes the impact of different attention mechanisms, such as HCGAM, FEECA, and FMSA, on the performance of the proposed model, XFloodNet. \cref{attention_wise} highlights the changes in performance when
attention mechanisms are individually or jointly omitted. The study evaluates configurations with one or more modules removed to assess the contribution of each attention mechanism. 
XFloodNet, incorporating attention mechanisms, achieved the highest performance, with an Acc of 96.01\%, Prec of 94.11\%, Rec of 92.56\%, and F1 of 93.33\%. These results underscore the effectiveness of integrating all attention modules, demonstrating their significant contributions to the model's performance.\\

XFloodNet (w/o HCGAM): Excluding HCGAM results in a drop in Acc to 94.13\% and F1 to 90.51\%, highlighting the critical role of this module in maintaining high feature alignment across modalities. The reduction in performance underscores the importance of HCGAM in enabling effective intra-modal and cross-modal interactions. Without this module, the model's ability to capture and integrate complex and hidden relationships diminishes, significantly impacting overall performance.  \\
{XFloodNet (w/o FEECA):} This version of XFloodNet shows a drop in performance across all the evaluation metrics. The Acc drops to 95.12\% and F1 reduces to 91.64\%, indicating its importance in the proposed approach. This shows that XFloodNet without FEECA is not able to capture spectral and channel-wise dependencies, weakening its feature representation and reducing overall performance. \\
XFloodNet (w/o FMSA): This variant of XFloodNet leads to an Acc drop of 94.92\% and an F1 reduction to 91.45\%. Without FMSA, the model's ability to extract spatial features across multiple scales diminishes, reducing its capacity to capture spectral as well as spatial dependencies and, thus, weakening overall feature representation. \\
XFloodNet (w/o (HCGAM and FEECA)): This type of XFloodNet shows a drop in Acc to 93.83\% and the F1 to 89.59\%. Their combined absence significantly weakens the model's ability to align features across modalities and capture spectral-channel dependencies, leading to a notable decline in performance.\\
XFloodNet (w/o (HCGAM and FMSA)): This configuration of XFloodNet exhibits a reduction in Acc to 92.72\% and F1 to 90.65\%. Due to their absence, the model's ability to align features across modalities and extract multi-scale spatial features decreases, leading to a noticeable drop in performance.  \\
XFloodNet (w/o (FMSA and FEECA)): With an Acc of 92.90\% and an F1 of 90.55\%, the combined absence of these mechanisms highlights the impact on both channel and spatial feature alignments, reducing the model’s ability to effectively capture and integrate multi-scale spatial and channel-wise dependencies. \\
In conclusion, each attention mechanism plays a crucial role in enhancing model performance. XFloodNet achieves optimal performance by leveraging the combined benefits of these modules, highlighting their importance for effective multimodal feature extraction and alignment in flood classification tasks.
\subsubsection{Qualitative Analysis}

\begin{table*}[pos=ht]
\centering
\begin{tabular}{lcccccc}
\toprule
 & \textbf{Sample 1} & \textbf{Sample 2} & \textbf{Sample 3} & \textbf{Sample 4} & \textbf{Sample 5} \\
\midrule
\textbf{Sample Details} & \begin{minipage}[t]{0.15\textwidth}\centering \includegraphics[width=\textwidth]{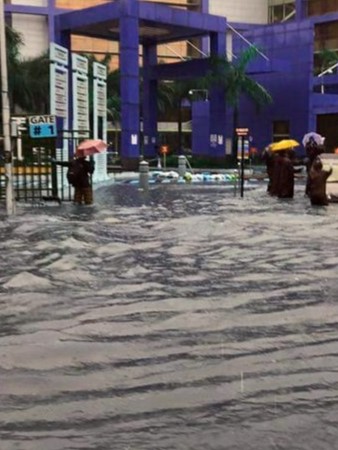}\\ {People are standing in the water.}\\
\textbf{Ground Truth: \textcolor{blue}{Flooded}}\end{minipage} & 
 
 \begin{minipage}[t]{0.15\textwidth}\centering \includegraphics[width=\textwidth]{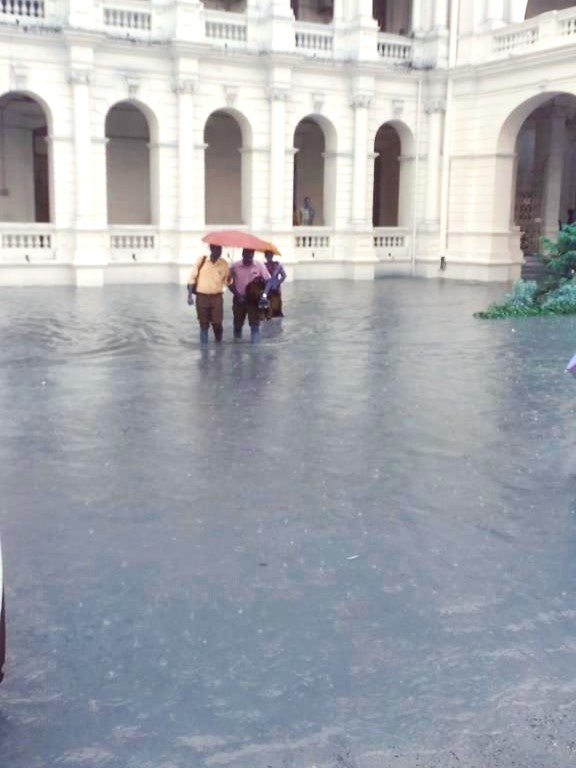}\\ {People walking in the flooded area.}\\ \vspace{0.4cm} \textbf{Ground Truth:  \textcolor{blue}{Flooded}}\end{minipage} & 
 
 \begin{minipage}[t]{0.15\textwidth}\centering\includegraphics[width=\textwidth]{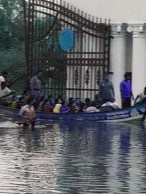}\\  The people are on a boat in the flooded area.\\ \textbf{Ground Truth:  \textcolor{blue}{Flooded}}  \end{minipage} &
 
 \begin{minipage}[t]{0.15\textwidth}\centering \includegraphics[width=\textwidth]{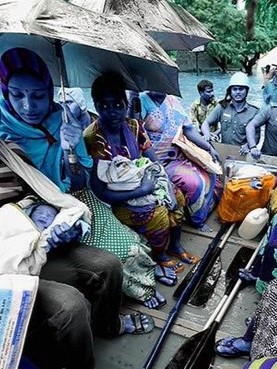}\\ People are being rescued via boat during floods. \\ \textbf{Ground Truth:  \textcolor{blue}{Non-Flooded}} \end{minipage} &
 
 \begin{minipage}[t]{0.15\textwidth}\centering \includegraphics[width=\textwidth]{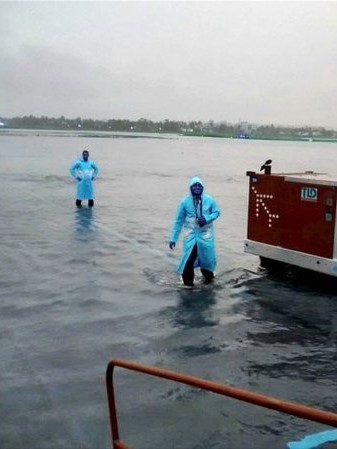}\\ The men are wearing blue suits in the water. \\  \textbf{Ground Truth:  \textcolor{blue}{Non-Flooded}} \end{minipage} \\
\midrule
\multirow{2}{*}{\textbf{Methods}}&\multicolumn{5}{c}{\textbf{Predictions}} \\
\cmidrule{2-6}
& \textbf{Post 1} & \textbf{Post 2} & \textbf{Post 3} & \textbf{Post 4} & \textbf{Post 5} \\ 
\midrule

\textbf{DIF-CAMM} &\textcolor{blue}{$$\cmark$$} &\textcolor{red}{$$\xmark$$}  &\textcolor{red}{$$\xmark$$}   &\textcolor{blue}{$$\cmark$$}  &\textcolor{red}{$$\xmark$$}  \\

\textbf{UAV-CNN} &\textcolor{red}{$$\xmark$$}  &\textcolor{red}{$$\xmark$$} &\textcolor{red}{$$\xmark$$}&\textcolor{blue}{$$\cmark$$}&\textcolor{red}{$$\xmark$$}  \\

\textbf{FloodVisionNet} &\textcolor{red}{$$\xmark$$}  &\textcolor{red}{$$\xmark$$}  &\textcolor{red}{$$\xmark$$}  &\textcolor{blue}{$$\cmark$$} &\textcolor{red}{$$\xmark$$}  \\

\textbf{AdaptNet} &\textcolor{red}{$$\xmark$$} &\textcolor{red}{$$\xmark$$}  &\textcolor{red}{$$\xmark$$}  &\textcolor{blue}{$$\cmark$$}  &\textcolor{red}{$$\xmark$$}\\

\textbf{CAMM} & \textcolor{blue}{$$\cmark$$} &\textcolor{blue}{$$\cmark$$}  &\textcolor{red}{$$\xmark$$} &\textcolor{red}{$$\xmark$$} &\textcolor{red}{$$\xmark$$}  \\

\textbf{DMCC} &\textcolor{blue}{$$\cmark$$} &\textcolor{blue}{$$\cmark$$}   &\textcolor{red}{$$\xmark$$}  &\textcolor{red}{$$\xmark$$}  &\textcolor{red}{$$\xmark$$}  \\

\textbf{RobertaMFT} &\textcolor{blue}{$$\cmark$$}  &\textcolor{blue}{$$\cmark$$}  &\textcolor{red}{$$\xmark$$}  &\textcolor{red}{$$\xmark$$}  &\textcolor{blue}{$$\cmark$$}  \\

\textbf{XFloodNet} &\textcolor{blue}{$$\cmark$$}  & \textcolor{blue}{$$\cmark$$} &\textcolor{blue}{$$\cmark$$}  &\textcolor{red}{$$\xmark$$}  &\textcolor{blue}{$$\cmark$$}  \\
\bottomrule
\end{tabular}
\caption{This table presents a qualitative analysis of XFloodNet's predictions. Correct predictions are indicated by \textcolor{blue}{\cmark}, while incorrect predictions are denoted by \textcolor{red}{\xmark}.}
\label{QualitativeAnalysis}
\end{table*}
This section presents a qualitative evaluation of state-of-the-art methods for flood classification, including the proposed method. The analysis is conducted using some samples from the test set, focusing on the performance of vision methods, multimodal methods, and XFloodNet, across several posts from the Chennai Flood dataset. The performance of each method is visually depicted using blue ticks (\textcolor{blue}{\cmark}) to signify correct predictions and red crosses (\textcolor{red}{\xmark}) to represent incorrect predictions or misclassifications. This qualitative assessment highlights the relative strengths and limitations of the evaluated methods, providing detailed insights into their efficacy under varying flood-related conditions. Annotator-provided or original labels serve as the ground truth for comparison. Representative posts, along with their ground truth labels and predictions, are depicted in \autoref{QualitativeAnalysis}, where Posts 1, 2, and 3 are categorized as ``Flooded'' and Posts 4 and 5 as ``Non-Flooded.''

The proposed framework, XFloodNet, exhibits superior classification performance across most test cases, which is attributed to its rich architectural design. Specifically, the Multimodal Feature Interaction Module facilitates effective intra-modal and cross-modal feature interaction and fusion; the Heterogeneous Convolutional Adaptive Multi-scale Attention Module enhances localized multi-scale feature extraction and global interactions via frequency features. The Cascading Convolutional Transformer Feature Refinement Module refines features through iterative processing and reverse feature harmonization technique. Collectively, these components enable XFloodNet to classify multimodal flood-related content, achieving state-of-the-art performance accurately.\\
Post 4 posed significant challenges for all evaluated multimodal methods, including XFloodNet, resulting in incorrect predictions. This failure is attributed to a discrepancy between the textual and visual modalities. While the accompanying text describes a flood rescue scenario involving a boat, the image portrays individuals seated in a boat holding umbrellas, which does not explicitly indicate an active flood event. This semantic incongruity between the modalities creates ambiguity, complicating accurate classification. Additionally, the presence of flood-related terms such as ``floods'' and ``rescued'' in the text introduced a bias that misled the model, causing it to prioritize textual features over contradictory visual evidence.\\
XFloodNet demonstrates exceptional performance across the majority of evaluated posts, underscoring its capability to integrate and analyze multimodal data effectively. By leveraging its rich architecture, the framework enables accurate classification even in complex scenarios, establishing XFloodNet as a state-of-the-art solution for flood-related multimodal content analysis.
\subsubsection{Computational Complexity}
\begin{figure*}
    \centering
    \includegraphics[width=1\linewidth]{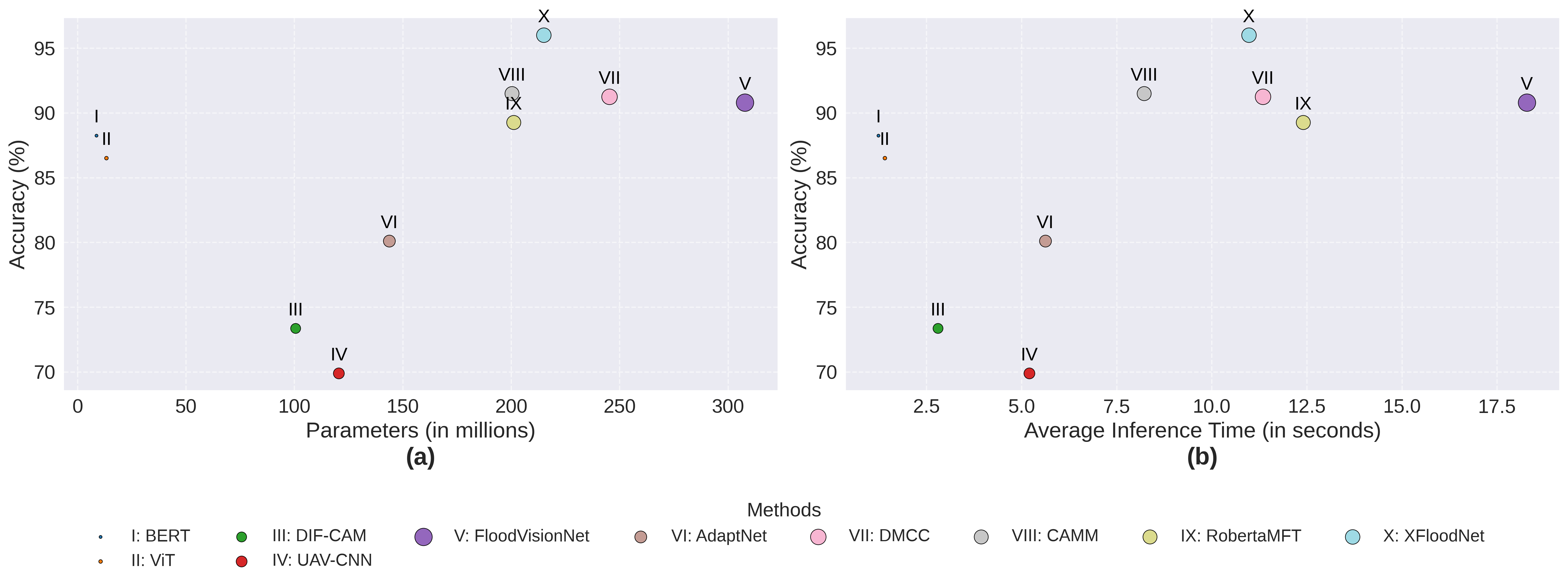}
    \caption{The figure illustrates the trade-offs between parameters (in millions), average inference time (in seconds), and accuracy (\%) for state-of-the-art methods. The size of each circle represents the parameter count, highlighting the computational demands and efficiency of each method. }
    \label{CC}
\end{figure*}

This section presents a comparative analysis of state-of-the-art methods,  visualizing the trade-off between predictive performance, model parameters, and computational cost. The analysis employs scatter plots to depict the relationships between model accuracy and two key metrics: the number of model parameters and average inference time. We have not taken into account the number of parameters of pre-trained models employed in all the state-of-the-art methods. The plot on \cref{CC}(a) illustrates the relationship between model parameters and accuracy. The x-axis represents the number of parameters in millions, and the y-axis represents the classification accuracy.
The plot on \cref{CC}(b) examines the relationship between accuracy and average inference time. The x-axis represents the average inference time in seconds, measuring the time required for the model to process a single input and generate a prediction. The y-axis, consistent with the left subplot, represents model accuracy. Each model is represented by a distinctively colored circular marker (bubble) in both subplots, with the color mapping to the model name as indicated in the legend. Roman numerals (I-X) within each marker provide a compact model identifier.\\
There is a general correlation between model size (number of parameters) and accuracy, as larger models can capture more complex data representations, enhancing predictive performance. However, this comes with increased memory demands and longer training times. The marker distribution underscores variations in parameter efficiency across models. The distribution of markers highlights the varying levels of parameter efficiency among the models. Some models achieve comparable accuracy with a significantly smaller parameter count, indicating more efficient architectures or optimization strategies.  Generally, models exhibiting higher accuracy also demonstrate longer inference times. This is attributable to the increased computational operations required by larger and more complex models. However, the plot also reveals variations in computational efficiency. Some models achieve similar accuracy with different inference times.
The figure shows an efficiency curve, highlighting models that balance accuracy with size or inference time. Models closer to the ``top-left'' of each subplot are optimal choices. The choice of model depends on the application: in resource-limited or real-time situations, smaller and faster models with slight accuracy reductions work best, while accuracy-focused tasks can use larger, slower models if resources are available.
\begin{table*} 
    \centering
    \setlength{\tabcolsep}{10pt} 
    \renewcommand{\arraystretch}{1.6} 
    \begin{tabular}{ c c c }
        \toprule
        \textbf{Sample} & \textbf{Text Modality} & \textbf{Image Modality} \\ \midrule
        Sample 1 & 
        \includegraphics[width=260px,height=110px]{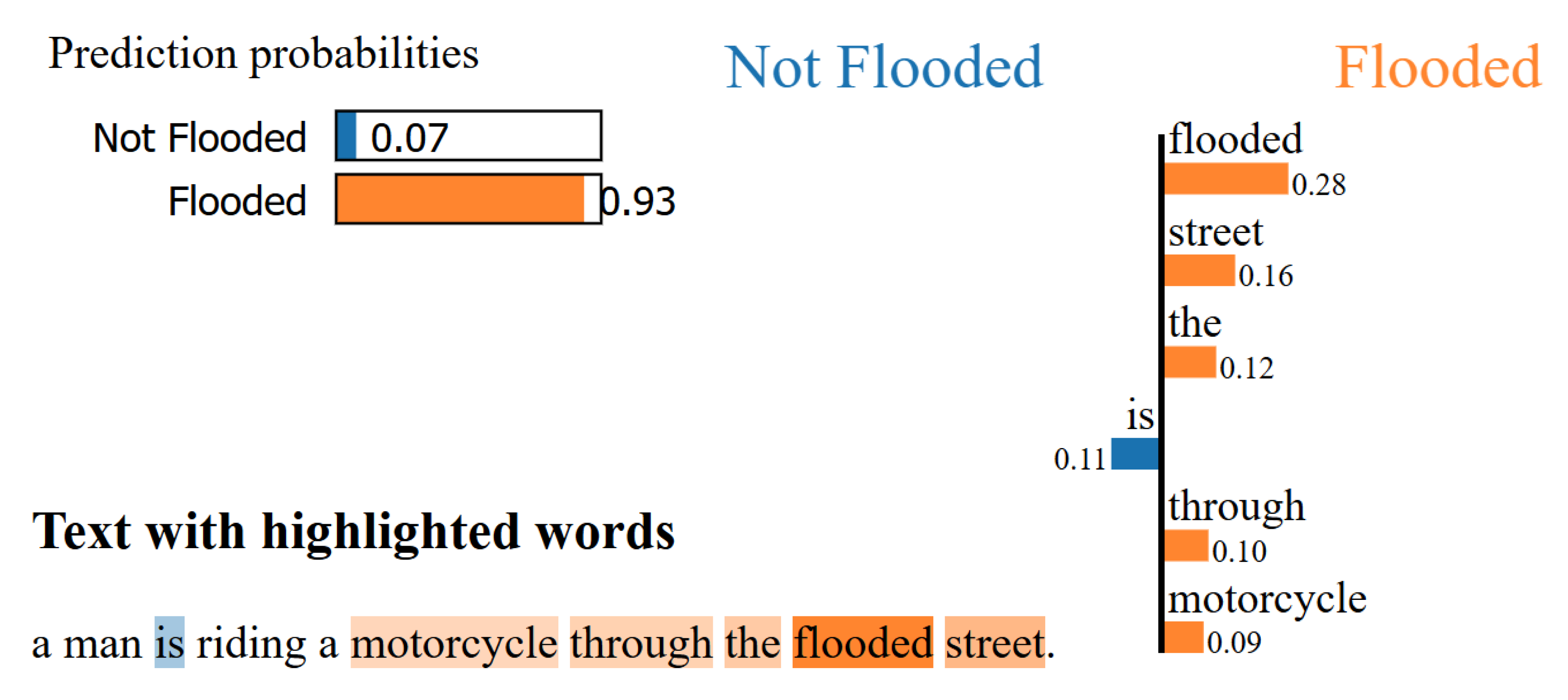} & 
        \includegraphics[width=130px,height=110px]{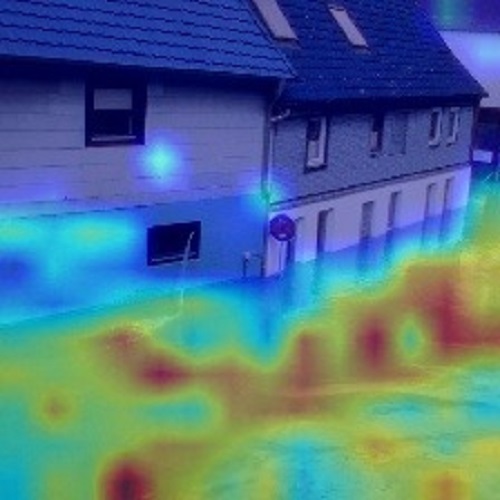} \\ \midrule
        Sample 2 & 
        \includegraphics[width=260px,height=110px]{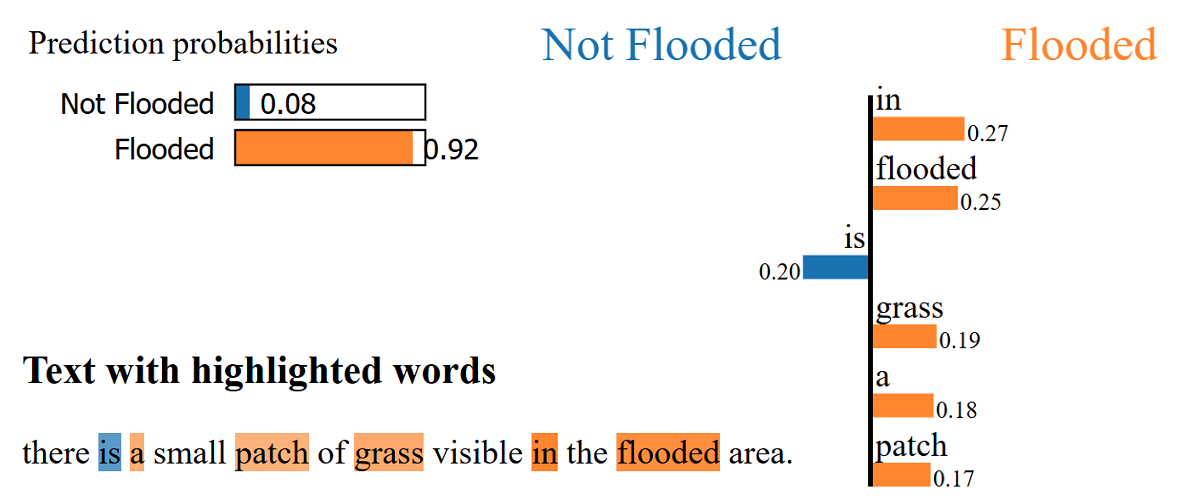} & 
        \includegraphics[width=130px,height=110px]{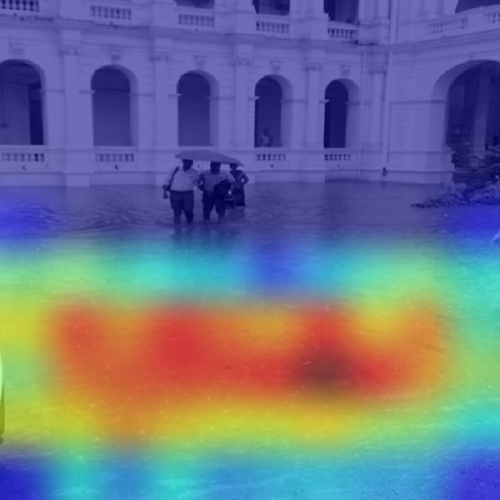} \\ \midrule
        Sample 3 & 
        \includegraphics[width=260px,height=110px]{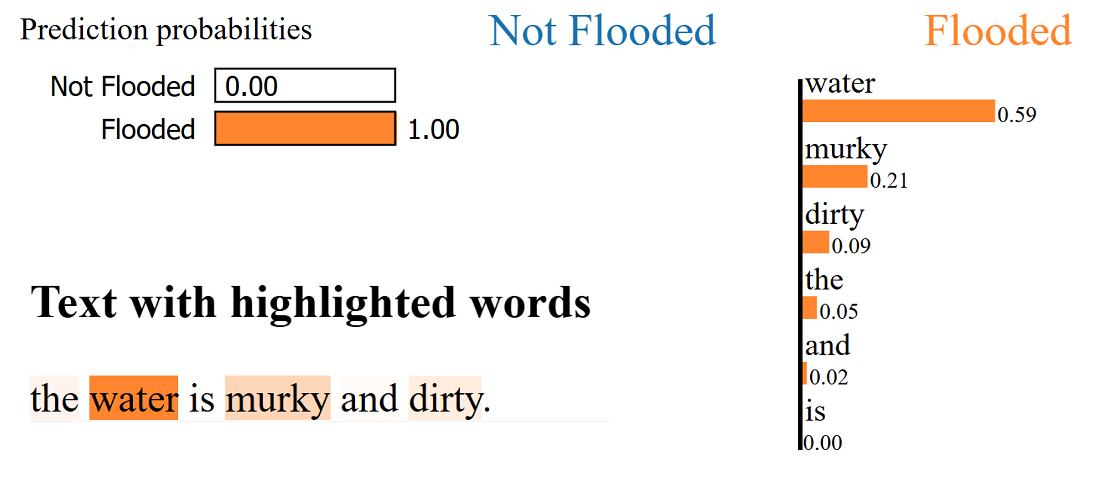} & 
        \includegraphics[width=130px,height=110px]{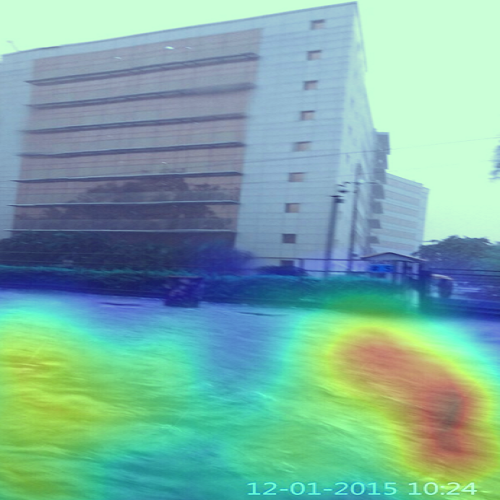} \\ \bottomrule
    \end{tabular}
    \caption{ Illustration of XFloodNet’s explainability framework: (a) LIME-based explanations for textual data, highlighting key terms contributing to the classification decision, such as `flood,' `water,' `murky,' and `dirty'; (b) Grad-CAM visualizations for image-based predictions, emphasizing critical regions influencing the model's classification. These explanations reveal patterns in the model's decision-making process, enhancing transparency and trust in disaster response applications.}
    \label{explain}
\end{table*}
\subsubsection{Explainability of XFloodNet}
XFloodNet achieves state-of-the-art performance in flood classification, demonstrating high accuracy and computational efficiency in disaster response applications. However, any model's adoption in critical scenarios requires a strong framework for explainability to ensure trust and transparency. To achieve this, XFloodNet incorporates LIME (Local Interpretable Model-Agnostic Explanations) \citep{lime} and Gradient-weighted Class Activation Mapping (Grad-CAM) \citep{gradcam} to provide clear insights into the reasoning behind its predictions for textual and visual modalities. LIME enables the interpretability of text-based predictions by highlighting the contribution of specific tokens or phrases. Grad-CAM, on the other hand, provides a visual explanation for image-based predictions by localizing critical regions within the input image that drive the model's decision. Together, these techniques establish a comprehensive explainability framework, offering insights into the interplay between textual and visual modalities.
By systematically analyzing LIME explanations across textual samples and Grad-CAM visualizations across multiple images, patterns in XFloodNet’s decision-making process can be identified. These patterns provide actionable feedback for improving the model's design in subsequent iterations. As demonstrated in \cref{explain}, such insights are crucial for fine-tuning the model’s multimodal representations and ensuring robust performance under varying real-world conditions.
This explainability framework not only enhances the model's transparency but also fosters trust among domain experts and stakeholders in disaster response. By bridging the gap between model outputs and human interpretability, XFloodNet facilitates informed decision-making and seamless integration into operational workflows, thereby advancing its applicability in critical disaster management scenarios.

\subsubsection{Statistical Analysis}
\begin{figure*}
    \centering
    \includegraphics[width=\linewidth]{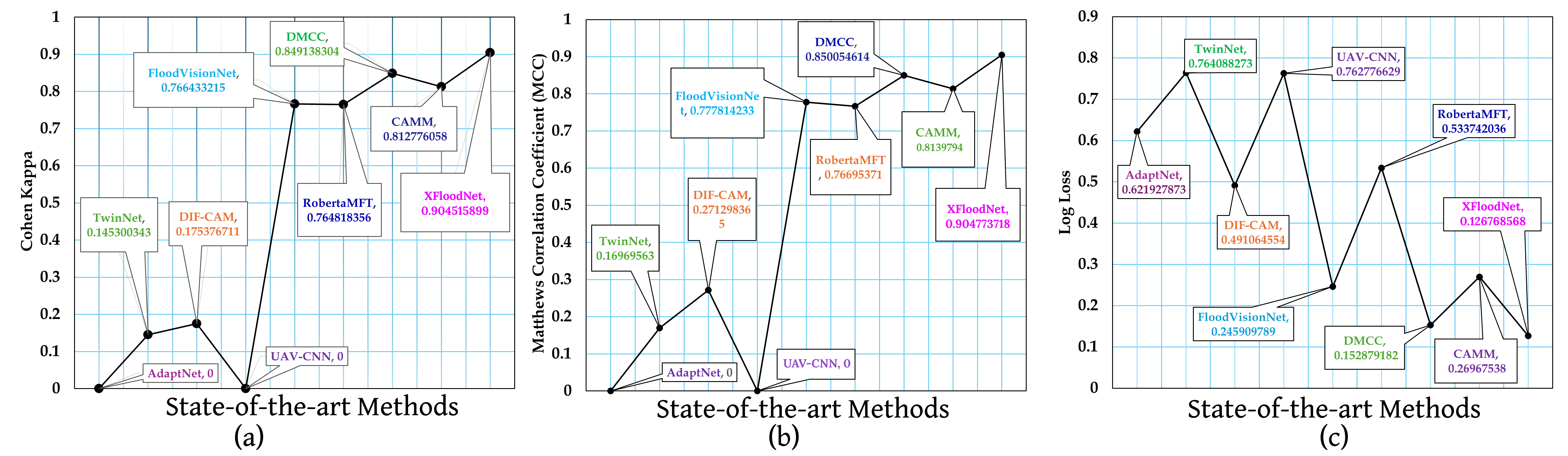}
    \caption{Evaluation of model performance using key statistical metrics: (a) Cohen's Kappa, illustrating the level of agreement between predicted and true labels; (b) Matthews Correlation Coefficient, providing a balanced measure of classification quality across all classes; and (c) Log Loss, assessing the accuracy of probabilistic predictions and their confidence levels.}
    \label{stats}
\end{figure*}

XFloodNet's performance was rigorously evaluated using a comprehensive suite of statistical tests and metrics. 
Beyond standard evaluation metrics such as accuracy, precision, recall, and F1-score, other metrics like Log Loss, Matthews Correlation Coefficient (MCC), and Cohen's Kappa were employed to provide a holistic assessment of the model's effectiveness. The decision to incorporate these metrics was inspired by the approach outlined in \citep{stat1}, which emphasizes the importance of leveraging diverse evaluation metrics to comprehensively evaluate classification models. These metrics enabled a deeper understanding of XFloodNet's performance compared to existing methods.\\
Additionally, McNemar's test was conducted to evaluate conflicting pairs, focusing on cases where two methods disagreed in their binary classifications. The test results revealed highly significant differences in error rates between XFloodNet and the existing methods, with \(p < 0.001\) across all comparisons. These extremely low p-values indicate a negligible likelihood that the observed differences are due to random chance, providing strong evidence that XFloodNet's performance is statistically distinct. Specifically, these results highlight that XFloodNet consistently produces fewer misclassifications compared to other models, particularly in edge cases where errors are more likely to occur. By demonstrating significant improvements in handling disagreements, McNemar's test validates XFloodNet's superior classification capabilities and establishes its robustness as a state-of-the-art solution.\\
MCC measures the quality of binary classifications, ranging from -1 to +1, where 0 represents random guessing. Cohen's Kappa quantifies the agreement between a model's predictions and actual values while accounting for chance agreement. Its values range from -1 (indicating perfect disagreement) to 1 (indicating perfect agreement), with scores between 0.8 and 1 reflecting near-perfect alignment. As highlighted in \autoref{stats}, XFloodNet achieved higher MCC and Cohen's Kappa values, demonstrating strong alignment with ground truth labels. Notably, Cohen's Kappa values approached 1, signifying near-perfect agreement, while reduced Log Loss further reinforced the model's predictive reliability. \\
In conclusion, the statistical analysis of this study,  along with the McNemar's test (\(p < 0.001\)), confirms XFloodNet as a progressive solution for flood detection and strongly support its use in critical disaster management workflows. XFloodNet's outstanding performance, with high accuracy, reliability, and efficiency, demonstrates its significant potential for improving disaster response efforts.

\section{Discussion}\label{sec5}
The proposed framework, ``XFloodNet'' demonstrates significant performance improvements in flood classification, attributed to the contributions of its novel components: Multimodal Feature Interaction Module (MFIM), Heterogeneous Convolution Adaptive Multi-scale Attention Module (HCAMAM), Cascading Convolutional Transformer Feature Refinement Module (CCTFRM), and  Unified Feature Fusion Module (UFFM). Each module plays a distinct and critical role in addressing specific challenges of multimodal feature extraction, interaction, fusion, and classification.\\\\
\textbf{Impact of the MFIM:}
This module enables the generation of rich and informative text descriptions for input images. By effectively interacting and fusing these text-based descriptions with the image features, the model captures intra-modal and cross-modal relationships between textual and visual modalities, which enhances the contextual understanding of the model, leading to improved feature representation and enhanced flood classification performance.\\\\
\textbf{Impact of the HCAMAM:}
The HCAMAM significantly boosts model performance by integrating advanced components such as the Heterogeneous Residual Extraction Network, Frequency-Enhanced Efficient Channel Attention, and Frequency-Modulated Spatial Attention. This module effectively captures key spatial, channel, and frequency features, facilitating robust feature extraction and fusion. It is designed to capture both point-wise and group-wise pixel information, ranging from coarse to fine-grained details of the visual data. Additionally, it incorporates both local and global contextual information, enhancing the model's ability to address complex visual tasks. Ultimately, the unified representation provided by this module enables precise and reliable flood classification outcomes. \\\\
\textbf{Impact of the CCTFRM:}
The CCTFRM architecture employs gated convolutional networks followed by the transformer and cascading operations to enhance the representational hierarchy. The Reverse Feature Harmonizer allows the features to be tuned and integrated precisely with input visual features. This approach strengthens feature representations at multiple levels, capturing complicated and complex relationships in the features for effective classification performance.
 \\
In brief, XFloodNet robustly and efficiently classifies flood events from social media data, enabling effective disaster response. This results in faster evacuations, more efficient resource allocation, and informed decision-making during crisis events.
\subsection{Limitations and Future Work}
XFloodNet, while demonstrating significant promise, faces certain limitations that present opportunities for future research.\\
XFloodNet effectively mitigates the noise and other related problems associated with social media data through its robust feature interactions, incorporation of frequency features, integration of local and global context, multimodal nature, and advanced attention mechanisms. However, further research can explore more sophisticated data preprocessing techniques, such as self-supervised learning and domain adaptation, to enhance data quality and improve model robustness.\\
 While XFloodNet demonstrates strong performance across diverse scenarios, further research is needed to enhance its generalizability across different geographical regions and temporal patterns. Exploring techniques such as multi-task learning can improve the model's ability to adapt to new and unseen flood events.\\
While this research focuses on flood classification, the core principles of XFloodNet can be extended to other disaster scenarios, such as wildfire detection, earthquake response, and hurricane monitoring. Further research can explore the applicability of this framework to a wider range of disaster management tasks.

\section{Conclusion}\label{sec6}
This work introduces XFloodNet, a novel multimodal framework designed to address the complexities of flood classification by effectively leveraging the power of multimodal learning.
XFloodNet includes novel techniques such as Hierarchical Cross-Modal Gated Attention Mechanism, Frequency-Enhanced Efficient Channel Attention, and Frequency-Modulated Spatial Attention, which collectively enable the model to dynamically adapt to diverse spatial and semantic features. The Hierarchical Cross-Modal Gated Attention mechanism refines intra-modal and cross-modal interactions by learning hierarchical dependencies between visual and textual features, enabling selective filtering of relevant information and capturing complex relationships for robust flood classification. The Frequency-Enhanced Efficient Channel Attention, which integrates high and low-frequency features within a single-channel framework, enhances feature representation while ensuring computational efficiency. On the other hand, Frequency-Modulated Spatial Attention integrates multi-scale convolutional and frequency domain features to enhance the model's focus on spatially relevant regions and global patterns. Additionally, the Cascading Convolutional Transformer Feature Refinement and the Reverse Attention Harmonization techniques synergistically refine features through gated convolutional networks, transformers, gated subtraction, scaling operations, and dynamic fusion. This integration optimizes feature interactions, enhances information flow, and significantly boosts classification performance. Comprehensive experimental evaluations on flood datasets demonstrated the efficacy of XFloodNet in achieving state-of-the-art results, surpassing existing methodologies in primitive evaluation metrics. By effectively addressing the challenges associated with unimodal analysis, static rule-based systems, and conventional convolutional networks, XFloodNet sets a new benchmark for flood event classification.
\\\\
\textbf{Acknowledgement}\\
We extend our gratitude to the Young Faculty Research Catalysing Grant (YFRCG) scheme, an initiative by IIT Indore, for awarding a research grant to Dr. Nagendra Kumar under Project ID: IITI/YFRCG/2023-24/03.

\bibliographystyle{elsarticle-num}
\bibliography{biblio}

\end{document}